\documentclass[twoside,11pt]{article}

\usepackage{jmlr2e}
\usepackage{amsmath}
\usepackage{color}

\DeclareMathOperator*{\argmax}{arg\,max}
\usepackage{gregmath}
\usepackage{algorithm}
\usepackage{algcompatible}
\renewcommand{\algorithmiccomment}[1]{\bgroup\hfill//~#1\egroup}
\usepackage{multirow}
\usepackage{multicol}
\usepackage{booktabs}
\usepackage{framed}
\usepackage{siunitx}
\usepackage{caption}
\usepackage{subcaption}
\usepackage{diagbox}
\usepackage{rotating}
\usepackage[table]{xcolor}

\newcommand{\rebuttaltext}[1]{\textcolor{black}{{#1}}}

\ShortHeadings{ProtoShotXAI: Using Prototypical Few-Shot Architecture for Explainable AI}{Hess and Ditzler}

\firstpageno{1}

\begin{document}

\title{ProtoShotXAI: Using Prototypical Few-Shot Architecture for Explainable AI}

\author{\name Samuel Hess \email shess@arizona.edu \\
       \addr Department of Electrical and Computer Engineering\\
       University of Arizona\\
       Tucson, AZ 85721 USA \\
       \name Gregory Ditzler \email ditzler@rowan.edu \\
       \addr Department of Electrical and Computer Engineering\\
       Rowan University \\
       Glassboro, NJ 08028 USA
       }

\editor{}

\maketitle

\begin{abstract}
Unexplainable black-box models create scenarios where anomalies cause deleterious responses, thus creating unacceptable risks. These risks have motivated the field of eXplainable Artificial Intelligence (XAI) which improves trust by evaluating local interpretability in black-box neural networks. Unfortunately, the ground truth is unavailable for the model's decision, so evaluation is limited to qualitative assessment. Further, interpretability may lead to inaccurate conclusions about the model or a false sense of trust. We propose to improve XAI from the vantage point of the user's trust by exploring a black-box model's latent feature space. We present an approach, ProtoShotXAI, that uses a Prototypical few-shot network to explore the contrastive manifold between nonlinear features of different classes. A user explores the manifold by perturbing the input features of a query sample and recording the response for a subset of exemplars from any class. Our approach is the first locally interpretable XAI model that can be extended to, and demonstrated on, few-shot networks. We compare ProtoShotXAI to the state-of-the-art XAI approaches on MNIST, Omniglot, and ImageNet to demonstrate, both quantitatively and qualitatively, that ProtoShotXAI provides more flexibility for model exploration. Finally, ProtoShotXAI also demonstrates novel explainability and detectability on adversarial samples. 
\end{abstract}


\section{Introduction}

Large-scale neural networks have made promising strides to reduce the classification error over large volumes of data \citep{he2016deep, silver2016mastering, van2016wavenet}. For example, residual networks have surpassed human-level accuracy on ImageNet \citep{he2016deep, ILSVRC15}. Policy networks have defeated professional humans in the game of Go \citep{silver2016mastering}, and WaveNet can automatically generate natural human-sounding speech from text \citep{van2016wavenet}. Unfortunately, these black-box models often lack human interpretable reasoning for their inferences because of the millions of weights, biases, and nonlinear activation functions in their architectural complexity. One example in recent literature presented evidence that the top ImageNet classification models were biased on textures rather than the shape bias, demonstrating that prior explanations and interpretations of these model might be incorrect \citep{geirhos2018imagenettrained}. Although textures may be sufficient for some tasks, the unexpected exploitation of texture leaves the model vulnerable to undesirable predictions, especially with potential exposure to real-world data anomalies \citep{alcorn2019strike}.

The concern of deleterious predictions from black-box models has triggered research efforts in eXplainable AI (XAI) that can be broadly organized into (1) the development of explainable models that are comparable in prediction performance to their black-box counterparts, and (2) the development of techniques to derive explanations from existing black-box models \citep{bodria2021benchmarking,adadi2018peeking}. 
Due to additional constraints of human interpretability, explainable models often struggle to achieve comparable performance with respect to their black-box counterparts. 
Additionally, explainable models are designed for a specific task, which limits their ability to extend to other tasks or datasets. 
In contrast, techniques that derive explanations from black-box models are support tools that can provide expectations on prediction behavior. 
The scope of explanations are often limited under the assumption(s) of the model and are not easily human interpretable. 
For example, some XAI approaches explain black-box models by weighting the importance of pixels. 
Unfortunately, these pixel attribution weights are still ambiguous because they require a human interpreter to extrapolate higher-level meanings from the weights (e.g., large pixel weights around a cat's ears can indicate the shape of the ears are important high-level features). 
\rebuttaltext{As discussed by \cite{guidotti2021evaluating}, this extrapolation of meaning only measures the adherence of the weights to the preconceived perception of individuals and leads to one of the core challenges with explaining black-box models: {\em there is no ground truth for the explanation itself}.}
In the cat example, large pixel weights around the ears could indicate that the shape of the ear is important for classification; however, the pixel weights could also indicate that the texture of the ear is what is truly important. Unfortunately, both of these explanations are just convenient for a human interpreter, but not necessarily correct. There could be many underlying true reasons for the pixel weights around the cat's ears. For example, it could be the ratio of power levels between pixels, the color, or many other possibilities. The actual rationale for the pixel weight is unknown and a challenge in XAI. 

A common approach to explain a model when ground truth does not exist is to involve actual human assessment through a survey. For example, \cite{jeyakumar2020can} performed Mechanical Turk surveys to determine user preference for explanations of deep neural networks (DNN) for image, text, audio, and sensor classification. 
Their study evaluated various state-of-the-art XAI against their proposed explain-by-example approach (namely, ExMatchina) and showed that users preferred ExMatchina for most tasks (image, audio, and sensor). Although the study provided an interesting insight into human preference, there is a disconnect between a human's explanation preference and the decision of the black-box model. That is, the empirical evaluation of user preferences can exploit the user's confirmation bias (i.e., what the user already believes are important features).

Since there is no ground truth for the explanation of a black-box model's prediction and human assessment of feature importance is inherently biased, we propose in this work to refocus XAI towards tractable tools for model exploration and trust. That is, we have developed a novel algorithm that is more amenable to exploring a model's manifold space w.r.t input feature manipulations. Specifically, we present ProtoShotXAI, a flexible black-box exploration approach that takes advantage of a model's trained feature representation layer, the weights of the classification layer (if available), and a Prototypical few-shot architecture to cross-compare a given query sample to class prototypes. \rebuttaltext{ProtoShotXAI allows for human-in-the-loop to manipulate an image under test and/or the support set to explore the changes to the feature representation. This user involvement provides more flexibility in testing prior hypotheses and a general exploration of model behavior.}

\rebuttaltext{In this work, we evaluate our novel approach against many state-of-the-art model explanations, including Grad-CAM, SHAP, LIME, RISE, and ExMatchina. We demonstrate that our approach can produce attribution maps that are popular with the XAI research literature and achieves comparable qualitative performance. Further, we present several novel experiments that allow human-in-the-loop driven model exploration that are unique to our approach. Most notably, deep few-shot architectures require both a query sample and a set of support samples to evaluate w.r.t. the query, and to our knowledge, our approach is one of the first explanation methods to account for both the query and support set.} The primary contributions in this work can be summarized as:
\begin{itemize}
  \item \rebuttaltext{A novel few-shot architecture, namely ProtoShotXAI, is presented as an XAI approach that merges local interpretability and prototype methods. ProtoShotXAI leverages human-in-the-loop interaction to drive a more useful cross comparison of feature representations. Our method has a broader impact in testing different hypotheses from a model's decision.}
  \item ProtoShotXAI is model-free in the sense that our approach is not specific to a particular neural network. Hence, our approach has broader applicability than some of the existing works in XAI.
  \item This work presents one of the first approaches to bridge XAI and few-shot architectures to increase the interpretability of few-shot neural networks and conventional models.
  \item Development of a tractable approach for model exploration and trust for black-box models, providing more insightful interpretations of real-world datasets.
  \item Publicly available code \& data at \url{https://github.com/samuelhess/ProtoShotXAI/} which includes cross-comparisons between common XAI approaches with consistent representations as well as four unique experiments for XAI: adversarial MNIST, revolving six, Omniglot XAI, and a multi-class ImageNet example.
\end{itemize}

The remainder of this manuscript is organized as follows. Section \ref{related_works} provides a background of recent and related works on XAI and few-shot learning. Section \ref{approach} describes the ProtoShotXAI approach, optimization, and use for learning localized regions of importance in images that explain a model's decision from a feature representation. Section \ref{Experiments} presents experiments on three benchmark datasets against several state-of-the-art XAI approaches. Section \ref{Discussion} discusses the findings from our experiments and variants of ProtoShotXAI. Finally, Section \ref{Conclusions} highlights the contributions, broader impacts, summary of the results and avenues for future research.

\section{Related Work}
\label{related_works}

\subsection{eXplainable AI (XAI)}

XAI can be broadly partitioned into two main categories: (1) approaches that are inherently explainable, and (2) approaches that derive explanations from black-box models \citep{bodria2021benchmarking, adadi2018peeking}. In this work, we focus exclusively on explaining existing black-box models that can be applied to DNNs. Within these methods, related work falls into the subcategory of locally interpretable approaches. Popular approaches in this subcategory include saliency maps \citep{simonyan2013deep}, Grad-CAM++ \citep{chattopadhay2018grad}, LIME \citep{ribeiro2016should}, SHAP \citep{lundberg2017unified}, RISE \citep{petsiuk2018rise}, and Anchors \citep{ribeiro2018anchors}.

Saliency maps are visualized as heat maps that indicate feature importance and the maps are used to determine correlated areas in an image that are valuable to a model's prediction. Saliency maps are common amongst all the approaches mentioned above and originally evolved from literature in gradient-based methods. Similar to how a neural network trains, gradient methods take the derivative of an input w.r.t. an output class activation. The derivative provides a direction of feature manipulations that, when applied to the input features, would result in a positive activation gain for being ``more like'' the respective class. More formally, given a trained black-box classifier that transforms a $D$-dimensional sample $\mathbf{x} \in \mathbb{R}^D$ into an estimated vector of $C$ classification scores $\tilde{\mathbf{y}} \in \mathbb{R}^C$, the gradient map $\mathbf{z} \in \mathbb{R}^D$ is simply the partial derivative of the loss w.r.t. the input features $\zbf = \frac{\partial \ybf_c}{\partial \xbf}$. Unfortunately, unregularized gradient-based changes to the input often exploit the neural network's function by providing unique, unrealistic, and/or unexpected combinations of noisy gradients rather than providing clear insight into an input feature's value w.r.t the network's prediction. As a result, many variants of gradient-based approaches apply spatial averaging and other regularization techniques to constrain the gradients and reduce their noise. This regularization and smoothing helps to provide consistency to the scale of a feature in a spatial region which, in turn, improves human interpretability \citep{selvaraju2017grad, chattopadhay2018grad}.

In contrast to gradient-based methods, Local Interpretable Model-agnostic Explanations (LIME) adds small perturbations to a given sample's input features and creates a linear model between the perturbations and the output predictions \citep{ribeiro2016should}. Specifically, a sample $\mathbf{x}$ is transformed into a  dataset $X'=\{\mathbf{x}_{1}',\ldots,\mathbf{x}_{N}'\}$ where $\mathbf{x}_{i}'$ are $N$ perturbations of $\mathbf{x}$. The dataset is passed through the classifier to produce a dataset of input samples and classification score pairings $T=\{(\mathbf{x}_{1}',\mathbf{y}_{1}'),\ldots,(\mathbf{x}_{N}',\mathbf{y}_{N}')\}$. A linear model can then be defined as $y'_{i}(c) = \mathbf{z}(c) \cdot \mathbf{x}_{i}'$ where the linear coefficients $\mathbf{z}$ represent the map of feature importance for each class $c$ and can be learned using multinomial logistic regression.
In contrast to allowing gradients to dictate the direction of changes to the input, LIME effectively tests local input-output regions around a given sample more systematically. LIME assumes that, whereas the total input-output space is often highly complex and nonlinear, the small perturbation region around a given sample under test can likely be approximated as linear. This assumption, along with the significant variance of coefficients $\mathbf{z}$ w.r.t the choice of input perturbations $X'$, have spawned many variants of LIME \citep{shankaranarayana2019alime, zafar2019dlime, peltola2018local, elshawi2019ilime, bramhall2020qlime}. Of particular noteworthiness, the original authors of LIME also created an alternative rules-based approach called Anchors. In Anchors, the features of an input sample are again perturbed; however, unlike LIME, they are perturbed intelligently with a multi-armed bandit to produce a set of ``anchors'' that consistently hold for an output decision regardless of other (non-anchor) feature values. Instead of the feature attribution weights produced by other approaches, Anchors produce a set of logical feature boundaries for a decision.

\rebuttaltext{Randomize Input Sample for Explanation (RISE)  is an approach similar to LIME \citep{petsiuk2018rise}. RISE perturbs the input data with a binary mask. As in LIME, a sample $\mathbf{x}$ is transformed into a dataset $X'=\{\mathbf{x}_{1}',\ldots,\mathbf{x}_{N}'\}$ where $\mathbf{x}_{i}'$ are $N$ {\em{randomly masked}} representations of $\mathbf{x}$ (i.e., $\mathbf{x}_{i}' = \mathbf{x} \odot \mathbf{m}_{i}$ where $\odot$ is the Hadamard product and $\mathbf{m}_{i}$ is the binary mask). The masked dataset is passed through the classifier to produce a dataset of input samples and classification score pairings $T=\{(\mathbf{x}_{1}',\mathbf{y}_{1}'),\ldots,(\mathbf{x}_{N}',\mathbf{y}_{N}')\}$. Unlike LIME, however, the data are not used to train a regression model. Instead, masks are weighted by the score for the class of interest, and the cumulative results are normalized as defined by }
\begin{align*}
  \mathbf{z} = \frac{1}{\mathbb{E} \left[  \mathbf{m} \right] N} \sum_{i = 1}^{N} y'_{i}(c)\mbf_{i}\
\end{align*}
\rebuttaltext{where $\mathbb{E} \left[\mathbf{m}\right]$ is expected value across the masks and is set externally via a hyperparameter \citep{petsiuk2018rise}. Setting a value of 0.5 would indicate that a pixel in the mask is a 0 or 1 with equal probability. It is worth noting that LIME also provides counterfactuals (i.e., attributions which indicate negative weights on the class of interest), whereas RISE is strictly positive attributions and does not consider counterfactuals.}

Often contrasted with LIME and RISE is Shapley Additive Explanations (SHAP), which is an approach that focuses on the Shapley values from cooperative game theory. Shapley values define the importance of an input feature as the likelihood that the outcome would be different if a model was trained without that feature. Naturally, training numerous models with different features removed is often intractable, so the majority of SHAP approaches use clever ways to approximate Shapley values with different underlying assumptions on the type of data or model \citep{sundararajan2017axiomatic, smilkov2017smoothgrad, erion2021improving}. Notably, Grad-SHAP and DeepSHAP  (also known respectively as GradientExplainer and DeepExplainer in the SHAP code libraries) are the versions of SHAP specifically designed for DNNs that we evaluate in this work. Grad-SHAP use a subset of $N$ training samples for background reference $T=\{(\mathbf{x}_{1},{y}_{1}),\ldots,(\mathbf{x}_{N},{y}_{N})\}$ and define the explainable map as
\begin{align*}
  \mathbf{z} = \mathop{\mathbb{E}}_{\mathbf{x'} \sim T, \alpha \sim U(0,1)}  \left[  (\mathbf{x} - \mathbf{x}_{i}') \frac{\partial \fbf(\mathbf{x}_{i}' + \alpha (\mathbf{x} - \mathbf{x}_{i}'))}{\partial \mathbf{x}}
  \right]
\end{align*}
where $\mathbf{x}_{i}'$ is sampled from a subset of training samples, $\alpha$ is sampled from a uniform distribution, and the partial derivative is the gradient at a feature layer $\fbf$ of the network w.r.t the input. The function $\fbf(\mathbf{x}_{i}' + \alpha (\mathbf{x} - \mathbf{x}_{i}'))$ represents a random hybrid between the input features under test $\mathbf{x}$ and a sampled background $\mathbf{x}_{i}'$. $\fbf$ can be the output of any feature layer up until the classification layer; however, the feature map becomes spatially coarser as the convolutional layers become coarser (and deeper layers in a CNN are often coarser). This observation means that the selection of a layer towards the middle should be preferred over a layer towards the end of the network. 

Another related subcategory to our work are prototype methods. XAI, through prototypes, attempts to provide the user with a different but relatable sample (i.e., a ``prototype'') that is an archetypical representation. Many of these approaches compute a distance between the prototypes and other samples in either the feature space of the input training data or a high-level feature manifold of a trained black-box model \citep{bien2011prototype, kim2016examples, koh2017understanding, papernot2018deep, gurumoorthy2019efficient}. The distance metric is used to find a representative class sample that is close to the query sample. For example, ExMatchina uses the final feature layer (i.e., the layer next to the classification layer) to compute the cosine similarity between the query sample and training samples to find the $N$ closest matches within that multi-dimensional manifold. Formally, given a trained black-box model and a sample $\mathbf{x}$, the exemplary sample $\mathbf{x'}_{e}$ that best explains $\mathbf{x}$ is 
\begin{align*}
  \mathbf{x'}_{e} =
  \argmax_{\mathbf{x'} \in T}
  \frac{
    \fbf^\T(\x) \fbf(\x')
  }{
   \| \fbf(\x)\|_2^2 \,\cdot\, \| \fbf(\x')\|_2^2
  }
\end{align*}
where the $\mathbf{x'}$ is from the the set of $N$ training samples $T=\{(\mathbf{x}_{1},{y}_{1}),\ldots,(\mathbf{x}_{N},{y}_{N})\}$ and $\fbf(\cdot)$ is the $K$-dimensional feature representation of $\xbf$. Intuitively, if $\mathbf{x}$ is in the training set ($\mathbf{x} \in T$) then $\mathbf{x}_{e}'$ is itself (i.e., $\mathbf{x}_{e}' = \mathbf{x}$).

\subsection{Few-Shot Learning}

In addition to related work in XAI, ProtoShotXAI leverages network architectures developed in the few-shot learning literature. Few-shot is a subcategory of classification where the goal is to categorize (with arbitrarily high accuracy) a set of unlabeled query samples into specific classes after being presented with a set of labeled support samples that contain one (or a ``few'') exemplary sample(s) from each class. In contrast to conventional classification, where many training samples are available for each class, few-shot training datasets have a limited number of training samples per class. For example, the MNIST dataset has roughly 6,000 samples per class. In contrast, the comparable few-shot character dataset (known as Omniglot) has only 20 samples per class (Section \ref{Experiments} presents further details of these datasets). Additionally,  a benchmark evaluation of few-shot networks is performed on classes that are mutually exclusive from the training classes. Note that this is in contrast to  conventional classification tasks that are only evaluated on classes in the training set. A few-shot task requires that one (or a ``few'') exemplary samples of classes are provided during testing and conventional classification provides no such supplementary information.

The concept of few-shot has been around for decades in machine learning literature \citep{fink2004object,fei2006one}; however, many recent neural network developments towards few-shot have emerged in response to the creation of benchmark datasets. In response to the aforementioned Omniglot dataset, Siamese networks \citep{koch2015siamese} created an architecture of two identical networks (i.e., networks with tied weights) where two different samples are presented at the input of each network. During training, Siamese networks compute a Euclidean distance at the feature layer of the network before the final sigmoid activation. The final activation is zero if two samples are from the same class and one if they are different (similar to the signature verification approach presented in \cite{bromley1994signature}). The network was evaluated by randomly subsampling one exemplary sample from 20 random Omniglot classes and classifying the remainder of the samples from the 20 classes (known as 1-shot, 20-way evaluation). It achieved 93.4\% classification accuracy.

Subsequent neural network developments extended the capability from one-shot to few-shot, improved the classification accuracy, and expanded the application to additional datasets. Notably, the authors of Matching Networks developed a network training scheme called episodic training where subsets of the training data are repeatedly (and randomly) selected to mimic the few-shot evaluation \citep{vinyals2016matching}. Similar to Siamese networks, Matching Networks uses tied weights and computes a distance metric at the feature layer. Each exemplary sample produces a distance w.r.t the sample under test and a neural attention mechanism selects the most likely class. They demonstrated state-of-the-art classification performance on the Omniglot benchmark as well as the more complex miniImageNet dataset introduced by Ravi and Larochelle \citep{ravi2016optimization}. Most metric-based approaches continue to use the episodic training scheme, including Prototypical Networks. The authors of Prototypical Networks demonstrated that they could achieve even better performance than Matching Networks with a simplified architecture (similar to Siamese Networks), a Euclidean distance metric at the feature layer, and a softmax over the distances \citep{snell2017prototypical}. Many few-shot architectures have been developed since Prototypical Networks, including metric architectures \citep{ye2018deep, scott2018adapted}, generative architectures \citep{wang2018low, antoniou2017data}, meta-learning architectures \citep{yoon2018bayesian, finn2018probabilistic}, or some combination of these approaches; however, Chen et al. showed many of the few-shot methods  have comparable results when using the same neural network backbone \citep{chen2019closer}.

\section{The ProtoShotXAI Architecture}
\label{approach}

XAI ExMatchina uses a distance function at the feature layer which is similar to the distance metric used in few-shot algorithms such as Matching Networks and Prototypical Networks. The objective for both ExMatchina and the few-shot approaches is similar as well: to find exemplary sample(s) that are close in the network's feature manifold w.r.t the sample under test. The difference is that the few-shot architectures are designed to train a network for contrastive classification and ExMatchina is designed to explain an existing trained network. Specifically, few-shot trains a network to minimize (at the feature layer) the distance between a query sample and an exemplary prototype whereas ExMatchina simply finds (at the feature layer of a trained classification network) the nearest training sample w.r.t the query. ExMatchina does not use an exemplary prototype nor does it use information from the trained classification weights.

Our ProtoShotXAI method is an XAI architecture that uses both an exemplary prototype and the information from the classification weights (when available) to explain a trained model's decision. Specifically, ProtoShotXAI augments the feature layer with classification weights and computes a similarity score between an exemplary prototype and a query sample that needs explanation. Rather than finding the closest exemplary sample (as performed by ExMatchina), ProtoShotXAI computes an average exemplary prototype that has been shown to work well in few-shot learning research \citep{snell2017prototypical, vinyals2016matching}. When used in few-shot explanations, ProtoShotXAI computes the average prototype at the feature layer; however, when used with conventional classification networks, our ProtoShotXAI approach combines the feature layer with the weights of the classification layer (i.e., just prior to feature summation and activation applied at the last layer) which we demonstrate improves explainability over state-of-the-art XAI approaches (see Section \ref{Experiments}). We argue that, whereas methods like ExMatchina and Grad-SHAP remove layers of the classification network, ProtoShotXAI uses all the layers and is, therefore, more representative of the trained model and more applicable across different architectures.

After adding the few-shot architecture to a trained network, the similarity score can be used to provide explainable exemplary samples similar to ExMatchina. Additionally, the change in similarity score w.r.t. perturbations at the input provide insight into input feature importance and a feature attribution map can be produced (e.g., similar to the maps produced by LIME and SHAP). Unlike previous approaches, ProtoShotXAI provides the user with more flexibility to explore input feature changes and perturbations. For example, by using ProtoShotXAI on digits, the user can rotate a six and observe the change in similarity score w.r.t classes six and nine (see Section \ref{Experiments}). Further, unlike LIME (which also uses input perturbations), there is no separate model. The feature maps can be generated directly from the trained classification network. 

\rebuttaltext{An important distinction between ProtoShotXAI and ExMatchina is that ExMatchina is strictly an explain-by-example approach. ExMatchina does not provide the users with an analysis of the distance in the feature space as ProtoShotXAI can support. This opens up an important research question, namely, ``{\em What would happen if we modified ExMatchina to analyze distances like we've done in our ProtoShotXAI approach}?'' In this work, we make a modification to ExMatchina and entitled this new approach ExMatchina*. ExMatchina* combines our contributions with the original ExMatchina for comparison.}

\rebuttaltext{In Table \ref{tab:algo comparison}, we summarize the key distinctions between our algorithms (ExMatchina* and ProtoShotXAI), and state-of-the-art XAI. 
Algorithms that provide counterfactuals indicate negative correlations between an input and an expected classifier output, which are not included in Grad-CAM, RISE, or ExMatchina. 
Concerning attribution weights on images, only RISE, LIME, and ProtoShotXAI provide full resolution maps (i.e., one-weight-to-one-pixel). Many of the approaches in Table \ref{tab:algo comparison} can be applied to any DNN classifier architecture; however, only LIME is known to be broadly applicable outside of DNN models; It is worth noting that few-shot is a special type of contrastive network that requires separate support and query sets and ProtoShotXAI is, to our knowledge, the only XAI model able to account for both in the XAI output.}

\begin{sidewaystable}
\centering 
\begin{tabular}{l|cccccccc}
\hline 
\cellcolor{gray!25}  & \cellcolor{gray!25} Deep & \cellcolor{gray!25} Expected  & \cellcolor{gray!25} Grad- & \cellcolor{gray!25} RISE &\cellcolor{gray!25}  LIME &\cellcolor{gray!25}  ExMatchina &\cellcolor{gray!25}  ExMatchina* & \cellcolor{gray!25}  ProtoshotXAI \\
\cellcolor{gray!25}  & \cellcolor{gray!25}  SHAP &  \cellcolor{gray!25}  Gradients & \cellcolor{gray!25} CAM &  \cellcolor{gray!25} & \cellcolor{gray!25}  & \cellcolor{gray!25}  & \cellcolor{gray!25}  (Our Mod.) & \cellcolor{gray!25}  (Ours) \\
\hline 
\hline 
 \cellcolor{gray!15} Counterfactuals & \checkmark  &  \checkmark &  &  & \checkmark &  & \checkmark & \checkmark \\
 \hline 
 \cellcolor{gray!15} Pairwise Comparisons &   &   &  &  &  & \checkmark & \checkmark & \checkmark \\
 \hline 
 \cellcolor{gray!15} Directly Applicable to &   &   &  &  &  &  &  & \checkmark \\
 \cellcolor{gray!15} Deep Few-Shot Networks &   &   &  &  &  &  &  &  \\
 \hline 
 \cellcolor{gray!15} Full Resolution &   &   &  & \checkmark & \checkmark &  &  &  \checkmark \\
 \cellcolor{gray!15} Attribution Maps &   &   &  & &  &  &  &  \\
 \hline 
 \cellcolor{gray!15} Deterministic &  \checkmark & \checkmark  & \checkmark &  &   & \checkmark & \checkmark & \checkmark \\
 \hline
 \cellcolor{gray!15} Model Agnostic &   &   &  &  & \checkmark &  &  &  \\
 \hline 
 \cellcolor{gray!15} Agnostic to DNN &   &   &  & \checkmark & \checkmark & \checkmark & \checkmark & \checkmark \\
 \cellcolor{gray!15} Classifier Architecture &   &   &  &  &  &  &  &  \\
 \hline 
\end{tabular}
\caption{\rebuttaltext{Tablular comparison of our algorithms (modified ExMatchina* and ProtoshotXAI) w.r.t. state-of-the-art methods evaulated throughout this manuscript.}}
\label{tab:algo comparison}
\end{sidewaystable}

\begin{figure}
\centering 
\includegraphics[width=0.95\textwidth]{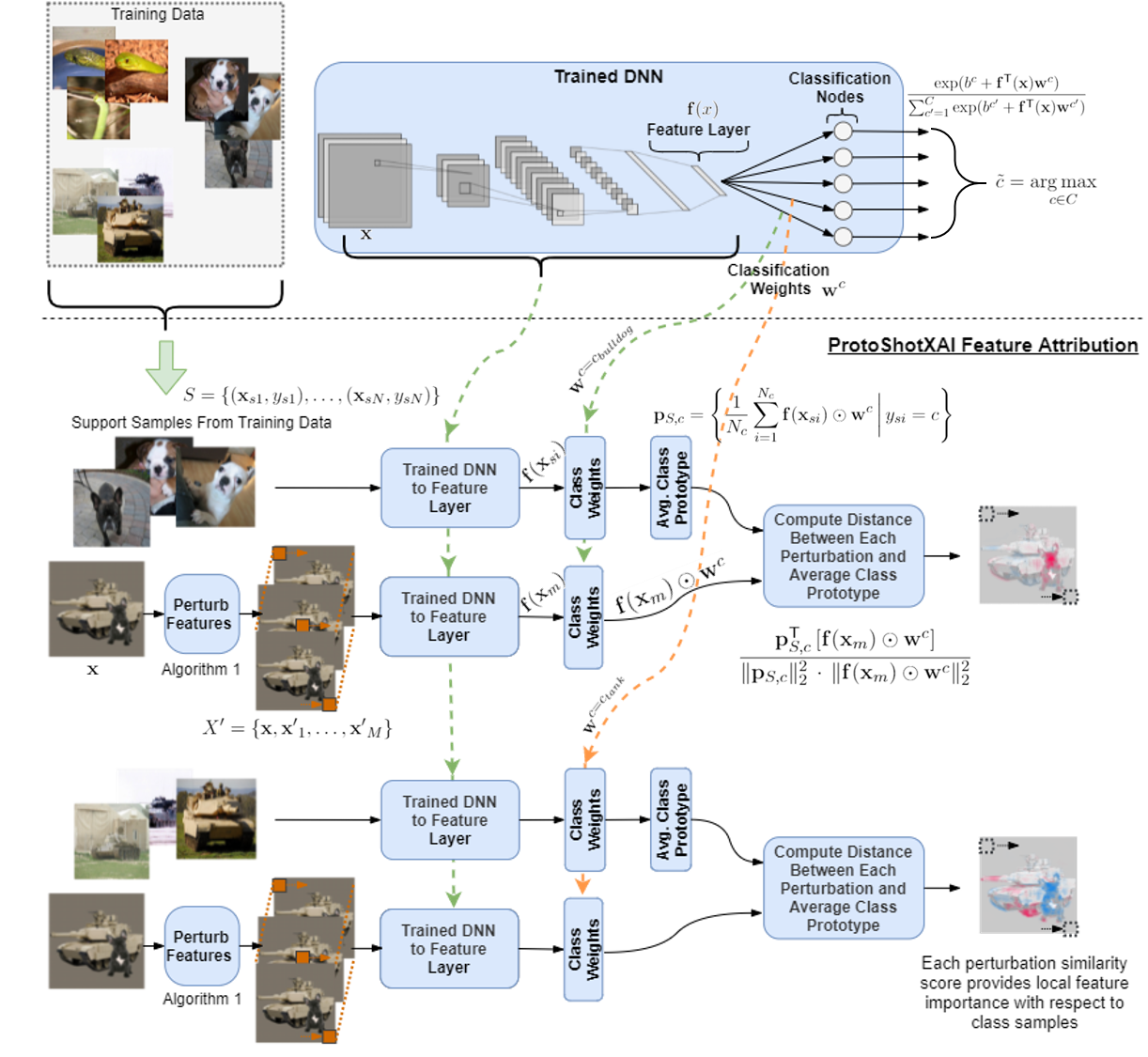}
\caption{
Overview of the architecture and feature attribution process of ProtoShotXAI. At the top of the figure is a trained DNN model and its corresponding training data. The ProtoShotXAI architecture has two branches. The first branch shows a set of support samples from the french bulldog passed through the trained DNN up to the feature layer and multiplied by the class weights of the french bulldog. A prototype vector is made from the average weighted features of the support samples. Similarly, the second branch contains the perturbations from the query sample (e.g., an image containing the tank and french bulldog). For each pixel perturbation image, a cosine distance is computed between the prototype vector and the weighted image features and the output is the feature attribution score at each pixel. The process done for the french bulldog support set is repeated for the tank support set to give positive and negative attribution scores w.r.t each class. 
}
\vspace{1in}
\label{fig:NetworkLayout}
\end{figure}

The entire process of constructing the ProtoShotXAI architecture and use it to create feature attribution maps is shown in Figure \ref{fig:NetworkLayout}. The ProtoShotXAI process starts with creating a trained classification network or using a preexisting trained network. Section \ref{Architecture} formally describes the process of constructing the ProtoShotXAI architecture from the baseline trained model and Section \ref{Implementation} describes one specific process for perturbing the data to produce feature attribution maps using ProtoShotXAI. The feature attribution maps (which can be directly compared against other approaches), as well as additional uses for ProtoShotXAI, are demonstrated in Section \ref{Experiments}.

\subsection{ProtoShotXAI Modification of a Trained Network}
\label{Architecture}

ProtoShotXAI modifies a pretrained classification network into the dual network architecture commonly used by many few-shot approaches such as Siamese networks, Matching networks, and Prototypical networks. That is, the trained DNN is replicated up to the feature layer for two branches: one branch contains the set of exemplary prototypes and the other branch contains variations of the query sample under observation. Formally, a classifier can be partitioned into a feature layer (the layer right before the classification layer) and a classification layer. The feature layer is represented by the nonlinear embedding function $\fbf(\cdot): \mathbb{R}^D \rightarrow \mathbb{R}^K$ which transforms a $D$-dimensional input sample $\mathbf{x} \in \mathbb{R}^D$ (such as an image) into a $K$-dimensional feature representation. At the classification layer, the features are multiplied by the trained static weights for each class node $\wbf^{c}$ where $c\in \{0,\ldots, C\}$ is a specific class out of the total number of class nodes $C$. The summation of the weighted features and bias $b^{c}$ go through a softmax activation function for each of the classes and the estimated class $\tilde{c}$ is the argmax of the activation as given by the following
\begin{equation}
  \tilde{c} = \argmax_{c \in [C]} 
  \frac{
    \exp(b^{c} + \fbf^\T(\xbf) \wbf^{c} )
  }{
    \sum_{c' = 1}^{C} \exp(b^{c'} + \fbf^\T(\xbf) \wbf^{c'} )
  }
\end{equation}
where $b^c \in \Rbb$ is a scalar that represents the bias. As shown in the bottom of Figure \ref{fig:NetworkLayout}, the first branch of the ProtoShotXAI architecture discards the activation, but keeps the trained network up until the feature layer and the weights of class. A subset $S=\{(\mathbf{x}_{s1},{y}_{s1}),\ldots,(\mathbf{x}_{sN},{y}_{sN})\}$ of support samples from a class are from the training dataset $Tr$ (e.g., images the french bulldog class). This support subset can be randomly selected (as done throughout this work) or, if the user knows a set of representative exemplary samples, they can be hand picked. All the $N$ $D$-dimensional support samples represented by $\mathbf{x}_{si} \in \mathbb{R}^D$ are passed through the DNN up until the feature layer and are multiplied by the weights of the specific class of interest (e.g., the trained weights from the french bulldog class). The prototype feature vector of the support set $\pbf_{S,c}$ is the average feature multiplied by the weights for class $c$.
\begin{equation}
\label{eq:Prototype_vector}
    \pbf_{S,c} = \left\{\frac{1}{N_c} \sum_{i = 1}^{N_c} \fbf({\xbf}_{si}) \odot \wbf^{c}\, \bigg|\, y_{si}=c\right\}
\end{equation}
where $\odot$ is the Hadamard product (i.e., an element by element multiplication operation). 
For the second branch of ProtoShotXAI, the sample under test $\mathbf{x}$ (represented as the image containing both the tank and french bulldog in Figure \ref{fig:NetworkLayout}) goes through a set of perturbations to observe the similarity response with respect to the prototype. Specifically, $\mathbf{x}$ is transformed into a perturbation dataset $X'=\{\mathbf{x}, \mathbf{x}_{1}',\ldots,\mathbf{x}_{M}'\}$. This perturbed dataset can easily be created by applying small changes to a different pixel (or set of pixels) in each sample. Similar to the support samples, each $\mathbf{x}_{m}'$ goes through the DNN until the feature layer and is then multiplied by the weights of the specific class; however, no average is taken for the perturbations. Instead, each perturbation is compared against the prototype using a cosine similarity metric as represented in the following equation:
\begin{equation}
\label{eq:ProtoShotXAI_vector}
  z_{m}(c,\xbf_{m}') = \frac{ \pbf_{S,c}^\T \left[ \fbf({\xbf}_{m}') \odot \wbf^{c} \right] } { \|\pbf_{S,c}\|_2^2 \,\cdot\, \| \fbf({\xbf}_{m}') \odot \wbf^{c}\|_2^2}
\end{equation}

\rebuttaltext{ProtoShotXAI differs from ExMatchina in three distinct ways which are key contributions of our work for XAI: (1) the ProtoShotXAI distance function uses a prototype of features for it's reference rather than the samples from a single feature, (2) ProtoShotXAI multiplies the features by the weights of a specific class node to get more representative features from the model, and (3) ProtoShotXAI uses input perturbations to cross-compare model responses to a baseline support set (like the feature attribution maps discussed in Section \ref{Implementation}). In the case of few-shot approaches that already have a dual network architecture, the class weights can be set to a vector of ones and the same XAI approach can be used, making ProtoShotXAI more consistent across architectures.}

\subsection{Using ProtoShotXAI to Produce Feature Attribution Maps}
\label{Implementation}

After the construction of ProtoShotXAI, the entire architecture can be used in many ways to explore the feature space of the classification network. The ProtoShotXAI architecture receives two sets of inputs. One set are the support input features to compare against (e.g., samples from a single class) and the other can be any set of feature alterations to a query sample under test, such as adding perturbations, skewing features, adding noise, and/or setting features to zero. The change in the similarity score provides a measure for the effects of the feature alterations w.r.t to the class. That is, changing features that are more important to the network model will have a larger effect on the change to the similarity score.

As an example, input feature attribution maps can be created similar to the ones produced by LIME, SHAP, and Grad-CAM for images. We illustrate this implementation in the bottom of Figure \ref{fig:NetworkLayout} and present it at as pseudo-code in Algorithm \ref{alg:ProtoShotXAI_Implementation}. The algorithm starts by choosing a set of input support image(s) (e.g., images of the french bulldog in Figure \ref{fig:NetworkLayout}) and image under test (e.g., an image containing both the tank and french bulldog in Figure \ref{fig:NetworkLayout}). A support set can be created from a random sample(s) of a class in the training data or any representative set of exemplary sample(s) and the image under test is any image to be evaluated w.r.t. the support. After the support and query data are selected, the algorithm calculates a baseline similarity metric $z_{ref}(c,\xbf)$ for reference. The algorithm then loops through each pixel in the image and replaces the RGB value of that pixel (or the small region surrounding that pixel) with an average value and computes the similarity metric for each spatial perturbation. Differences between the reference similarity score and the similarity score when the pixel is replaced are the feature attribution of that pixel. As the similarity score decreases, the pixel is weighted as being more important, and vice versa, as the similarity score increases.

\begin{algorithm*}[t] 
 \caption{Implementation of ProtoShotXAI to Create Feature Attribution Maps}
 \label{alg:ProtoShotXAI_Implementation}
 \begin{algorithmic}[1]
 \renewcommand{\algorithmicrequire}{\textbf{Input:}}
 \renewcommand{\algorithmicensure}{\textbf{Output:}}
 \REQUIRE{Any representative set of $N$ support image(s) $S=\{(\mathbf{x}_{s1}),\ldots,(\mathbf{x}_{sN})\}$ for a single class $c$, a image under test $\mathbf{x}$, average RGB pixel vector $\mathbf{R}_{RGB}$, a trained classification model with frozen parameters}
 \ENSURE{Feature attribution map $\mathbf{z}$.}
   \STATE 
   $z_{ref} = \frac{ \pbf_{S,c}^\T \left[ \fbf({\xbf}) \odot \wbf^{c} \right] } { \|\pbf_{S,c}\|_2^2 \,\cdot\, \| \fbf({\xbf}) \odot \wbf^{c}\|_2^2}$
   \COMMENT{\tt Reference score of $\mathbf{x}$ w.r.t $S$}
   \FOR[\tt Loop through columns of the image]{$i = 1,\ldots,I$}
   \FOR[\tt Loop through rows of the image]{$j = 1,\ldots,J$}
   \STATE ${m}= j + (i-1)J$ \COMMENT{\tt create a perturbation index for each 2D pixel}
   \STATE $\mathbf{x}_{m}' = \mathbf{x}$ \COMMENT{\tt $\mathbf{x}_{m}'$ is initialized as $\mathbf{x}$ before each perturbation}
   \STATE $\mathbf{x}_{m}'(i,j) = \mathbf{R}_{RGB}$ \COMMENT{\tt perturb the $i,j$th pixel to average RGB value}
   \STATE $z(i,j) = z_{ref} - 
   \frac{ \pbf_{S,c}^\T \left[ \fbf({\xbf}_{m}') \odot \wbf^{c} \right] } { \|\pbf_{S,c}\|_2^2 \,\cdot\, \| \fbf({\xbf}_{m}') \odot \wbf^{c}\|_2^2}
   $ 
   \COMMENT{\tt Score for $\mathbf{x}_{m}'$ w.r.t $S$}
   \ENDFOR
   \ENDFOR
 \end{algorithmic}
\end{algorithm*}

\section{Experiments}
\label{Experiments}

This section empirically demonstrates the proposed ProtoShotXAI approach on three benchmark datasets, \rebuttaltext{namely: MNIST, Omniglot, and miniImageNet. ProtoShotXAI is compared against several state-of-the-art XAI approaches, including ExMatchina, ExMatchina*, Grad-CAM, Grad-SHAP, RISE, and LIME}. We describe multiple experiments where ProtoShotXAI can be used to explore the original classification network qualitatively and quantitatively, and demonstrate how our approach achieves comparable and often superior results to more complex algorithms for XAI. Throughout this section, we use \textbf{boldface} text when referring to characters. \rebuttaltext{Specific details related to implementation, the datasets, and the neural network architectures is provided in Appendix \ref{app:Implementation_Details}}.

\subsection{MNIST}
In this section, we demonstrate three different aspects of ProtoShotXAI with experiments for the MNIST dataset. In the first experiment, we qualitatively evaluate the locally interpretable algorithms by visualizing the feature representations of the digit \textbf{eight}. \textbf{Eight} is a unique digit due to the subset of digits that are contained within it. For example, the digit \textbf{three} is commonly contained as a subset of \textbf{eight} as well as parts of \textbf{five}, \textbf{six}, and \textbf{nine}, making it convenient for cross-class explanations.  In the second experiment, we demonstrate how ProtoShotXAI can be used to quantitatively assess the classification change as a \textbf{six} is rotated until it is classified as a \textbf{nine}. Finally, the last MNIST experiment analyzes the XAI approaches on an adversarial dataset. The adversarial dataset is unique because the true class and the predicted class from the classifier are often different and the features of ProtoShotXAI provide insight into the network's classification errors.

\subsubsection{Feature Attribution on Eight}
\begin{figure}
\centering 
\includegraphics[width=1.0\textwidth]{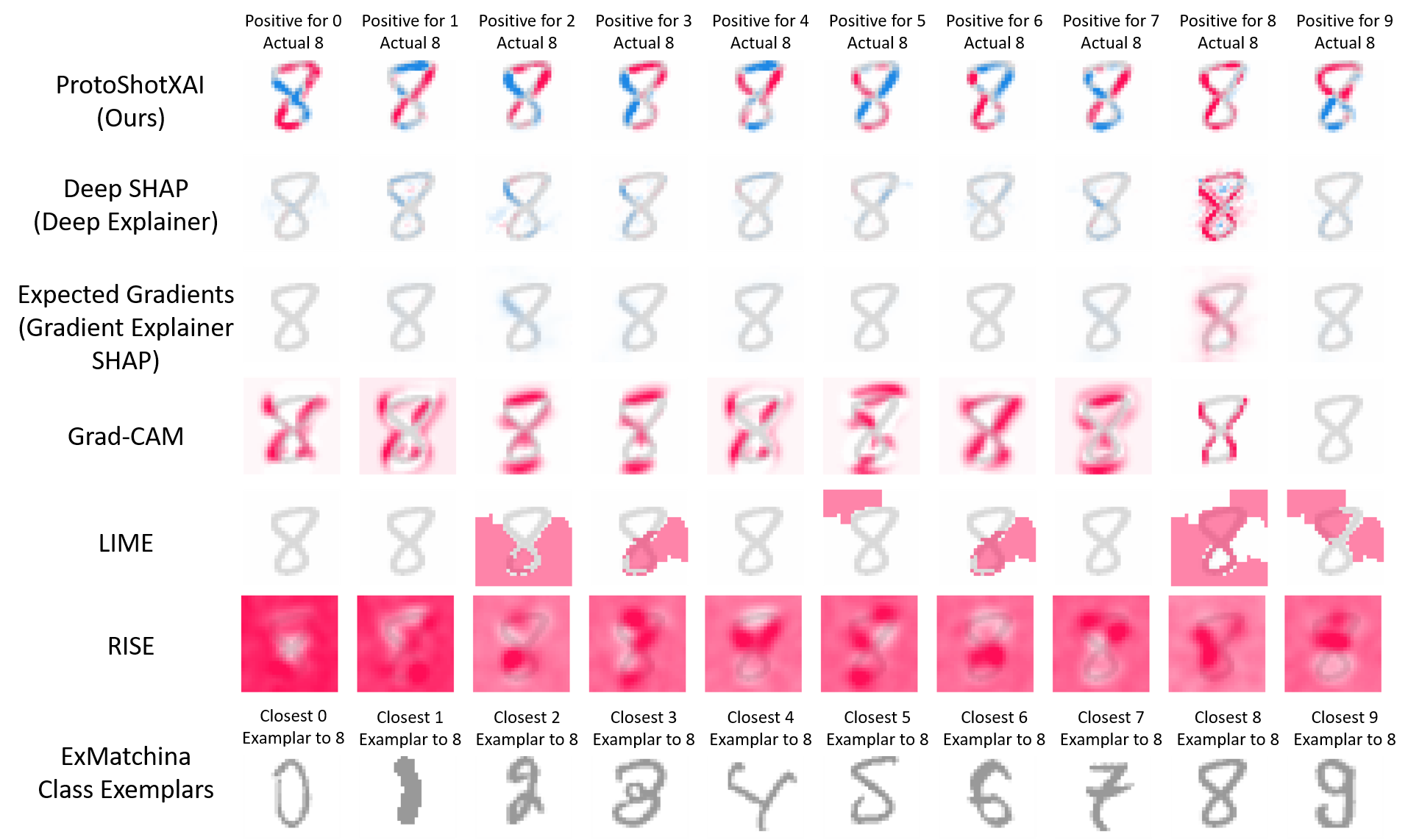}
\caption{Example of feature attribution weights of various approaches on an exemplary character \textbf{eight} (shown in gray throughout). Consistent with the SHAP conventions, red areas indicate positive feature attributions and blue areas indicate negative feature attributions. ProtoShotXAI shows more contrast in features (e.g., positive attributes for digits \textbf{two}, \textbf{three}, \textbf{five}, and \textbf{seven} as sub-features of the digit \textbf{eight}) compared to other methods. ExMatchina is the one prototype method shown due to the mathematical similarity to ProtoShotXAI. ExMatchina only provides the nearest exemplary samples in a class which is apparent from provided exemplary digits like \textbf{three}, \textbf{five}, and \textbf{nine} but does not provide the same analytical clarity of feature attributions.}
\label{fig:mnist_example}
\end{figure}

In the first experiment, we qualitatively evaluated ProtoShotXAI with an MNIST sample of the digit \textbf{eight}. The digit \textbf{eight} was chosen as a test sample due to the presence of multiple digits contained within the shape of the \textbf{eight}. 
The same \textbf{eight} was tested against each digit class (0-9) using the seven state-of-the-art XAI algorithms, namely, ProtoShotXAI, Deep SHAP, Grad-SHAP, Grad-CAM, LIME, RISE, and ExMatchina. The reference pixel value used in ProtoShotXAI is set to zero which indicates a lack of digit presence in a pixel. The results of each algorithm were normalized using the same standard applied in the SHAP work, which is necessary for a fair comparison, and the results are displayed in Figure \ref{fig:mnist_example}.

One of the first observations from Figure \ref{fig:mnist_example} is that pixels with negative feature attributions are much more prevalent in ProtoShotXAI over the other approaches. The SHAP variants are the only other methods with negative feature attributions; however, the magnitude of those attributions are relatively small. The prominent negative features of ProtoShotXAI are attributed to the weights of the classification nodes which cause strong negative responses to counterfactual features of the given class (see Section \ref{sec:adversarial} for more examples with adversarial perturbations). As a result, the remaining positive features in ProtoShotXAI highlight expected areas of interest. For example, the positive features for \textbf{two} are symmetrical to \textbf{five} about a vertical axis and the positive features for \textbf{six} are symmetrical to \textbf{nine} about a horizontal axis. Concerning the digit \textbf{three}, there is significantly more clarity in the ProtoShotXAI positive and negative features when compared to other methods. Additionally, ProtoShotXAI is more spatially precise than other methods. SHAP and Grad-CAM use only part of the convolution network layer and rely on spatial interpolation which causes a spatial blurring effect. LIME uses spatial segmentation areas that cause entire regions to be flagged as a positive or negative feature attribute. ExMatchina is the only explain-by-example approach shown due to its mathematical similarity to ProtoShotXAI but is not intended to provide cross-class comparisons. Hence, the closest samples ExMatchina provide are arguably ``\textbf{eight}-like'' (see exemplar \textbf{three}, \textbf{five}, and \textbf{nine}) but do not provide feature clarity like the other approaches in this context.

\subsubsection{Revolving Six}
We also performed a simple -- yet meaningful -- experiment that shows ProtoShotXAI can be used for more than feature attribution maps, the similarity score it produces can bring fidelity to classification. Further, ProtoShotXAI has better agreement than the class of the nearest exemplary sample (i.e., ExMatchina). To demonstrate these properties of ProtoShotXAI, we consider an image of the digit \textbf{six} that is rotated from 0$^{\circ}$  to 360$^{\circ}$. 
If a \textbf{six} is rotated 180$^{\circ}$ about the center of the digit, the image can be incorrectly classified as a \textbf{nine} by the network. It is worth mentioning that most humans would also incorrectly classify the rotated \textbf{six} as a \textbf{nine}. A question that arises from this scenario is: as a \textbf{six} is rotated from 0$^{\circ}$ and 180$^{\circ}$, when does this transition occur that causes a different classification, and how rapid is the transition? 

\begin{figure}
\centering 
\includegraphics[width=1.0\textwidth]{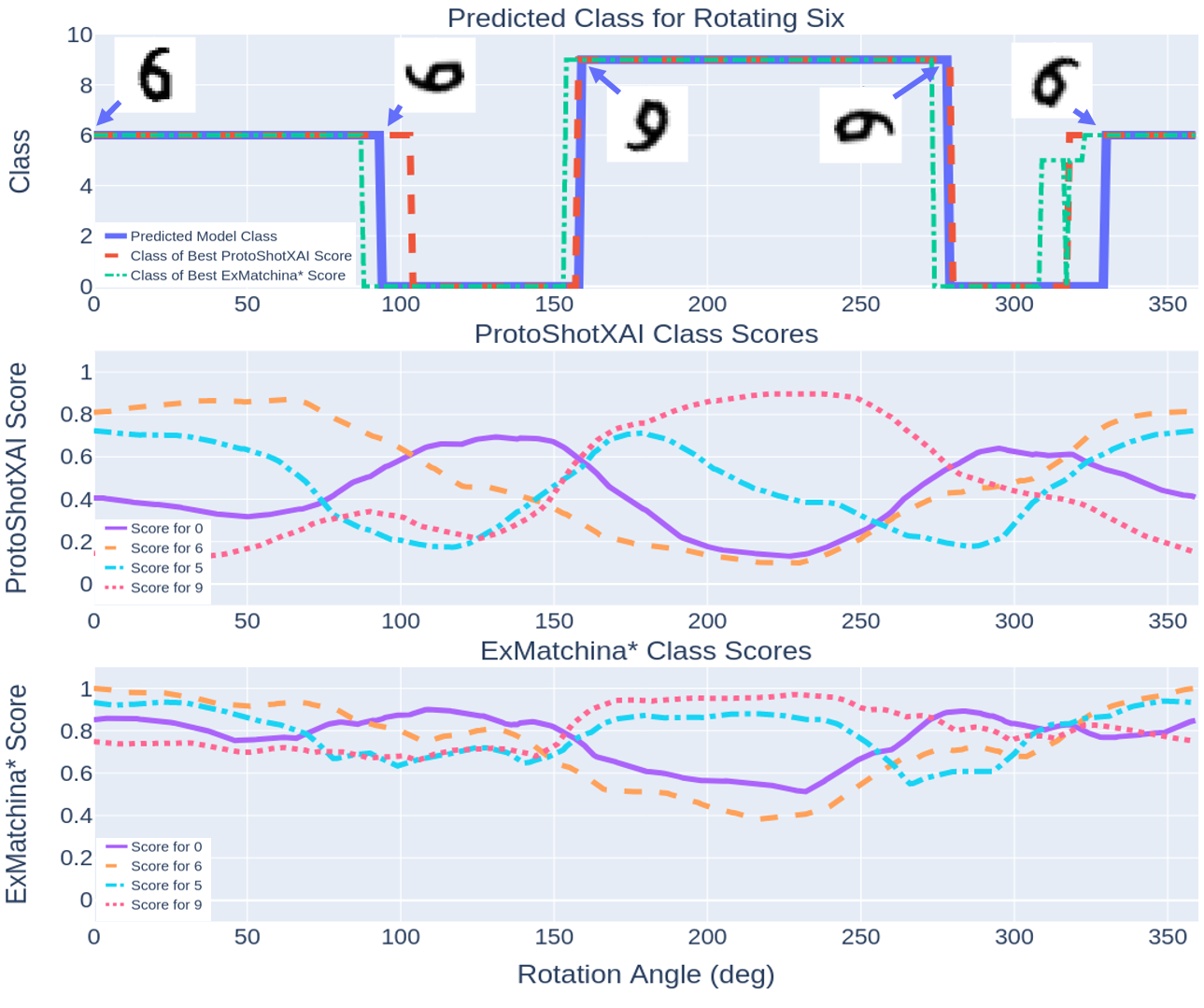}
\caption{Predicted class as a \textbf{six} is rotated from 0$^{\circ}$ to 359$^{\circ}$ (blue line in the top plot) and corresponding ProtoShotXAI and EXMatchina$^*$ scores for classes \textbf{zero}, \textbf{six}, and \textbf{nine}. The red and green lines in the top plot result from assigning the class of the largest similarity score at each rotation angle for ProtoShotXAI and ExMatchina$^*$, respectively. ProtoShotXAI has more consistency with the model predictions due to the prototype averaging.}
\label{fig:r6_example}
\end{figure}

In this experiment, we rotate a \textbf{six} counterclockwise from 0$^{\circ}$ to 359$^{\circ}$ in increments of 1$^{\circ}$ then pass each rotated sample to a DNN. 
The model classifying these different rotational views of the \textbf{six} only predicts three different classes throughout the rotation: 
a \textbf{six}, a \textbf{zero}, and a \textbf{nine}. The transitions occur at 93$^{\circ}$ (transition from 6 to 0), 158$^{\circ}$ (transition from 0 to 9), 278$^{\circ}$ (transition from 9 to 0), and 329$^{\circ}$ (transition from 0 to 6). 
The top plot in Figure \ref{fig:r6_example} shows the transition of the class predicted by the model (solid blue line), the ProtoShotXAI score (dashed red line), and the ExMatchina* score (dashed green line).   
For each of these rotated images, we use the ProtoShotXAI architecture to compute the distance between the rotated sample and a prototype of 100 random exemplary samples from each digit class of interest (i.e., classes \textbf{six}, \textbf{zero}, \textbf{five}, and \textbf{nine}). 
ProtoShotXAI scores for these classes of interest as the \textbf{six} is rotated are shown in the middle plot of Figure \ref{fig:r6_example}. For comparison, the same 100 random exemplary samples are used in ExMatchina* to retrieve the best cosine similarity within each class, which are shown in the bottom plot of Figure \ref{fig:r6_example}. 
Note that this evaluation of ExMatchina* was not proposed in the authors' original implementation. Since we revised ExMatchina from an explain-by-example approach to an analytical approach closer to ProtoShotXAI, we represent it as ExMatchina$^*$.

The ProtoShotXAI and ExMatchina$^*$ scores have similar behaviors that are correlated to the model's predictions; however, these scores have better explainable fidelity. 
For example, the scores from ProtoShotXAI show a transition between classes of approximately 20$^{\circ}$-30$^{\circ}$ before the model is certain about the next class. 
The difference between ProtoShotXAI and ExMatchina$^*$ in this synthetic example is that ProtoShotXAI finds the cosine similarity between the rotated \textbf{six} and the {\em average prototype} of the 100 exemplary samples, whereas ExMatchina$^*$ computes the cosine similarity between the rotated \textbf{six} and all the 100 exemplary samples then chooses the {\em prototype with the maximum similarity}. The result from ProtoShotXAI is more stable because the reference class prototypes will not change as the \textbf{six} rotates; however, this observation can not be said about the response of ExMatchina$^*$'s score. The stability of ProtoShotXAI is further demonstrated by comparing the class predicted by the model (shown in the top plot of Figure \ref{fig:r6_example}) with the class that corresponds to the maximum score for ProtoShotXAI (dashed red line) and the maximum cosine similarity used in Exmatchina$^*$ (dashed green line). The output of the predicted class from ProtoShotXAI scores is more consistent with the class predicted by the model whereas the ExMatchina$^*$ cosine distance produces more deviations and even temporarily predicts a \textbf{five}. Therefore, we conclude ProtoShotXAI produces a similarity score that brings explainable fidelity to a model's prediction and provides more insight than ExMatchina$^*$.

\subsubsection{Adversarial MNIST}
\label{sec:adversarial}

In the context of classification, adversarial data are crafted to exploit a model's behavior to produce incorrect predictions by adding a small perturbation to the original image that is imperceivable to a human but fools the model \citep{Goodfellow2014arxiv, Sadeghi2020TETCI, Qiu2021ToC}. For example, specific small perturbations to an MNIST digit can appear to be minor alterations to the human eye yet cause drastic changes to the output prediction of a trained neural network. This behavior causes hesitancy in integrating neural networks in precarious situations where an adversary might be able to inflict malicious data with the intent of causing deleterious predictions. As a result, a relevant goal of XAI is to explain when and why adversarial data corrupts model behavior. If XAI can explain when and why, then deleterious predictions from an adversary can be bounded or mitigated which increases user trust in the black-box model.

To explore the adversarial behavior of ProtoShotXAI, we crafted an adversarial dataset by taking the trained neural network from Table \ref{tab:mnist-model-summary} and training it on the MNIST dataset. The adversarial samples are generated from the testing dataset using the {\em Fast Gradient Sign Method} (FSGM), which generates adversarial samples of the form \citep{Goodfellow2014arxiv}: 
\begin{align}
    \x_{\text{adv}} = \x + \epsilon \, \text{sign}( \nabla_\x L(\theta, \x, y) )
\end{align}
where $\epsilon>0$ is the adversarial budget, $\theta$ are the parameters of the network and $J(\cdot, \cdot, \cdot)$ is the categorical cross-entropy loss. The adversarial samples are presented to the ProtoShotXAI network. 

\begin{figure}
\centering 
\includegraphics[width=1.0\textwidth]{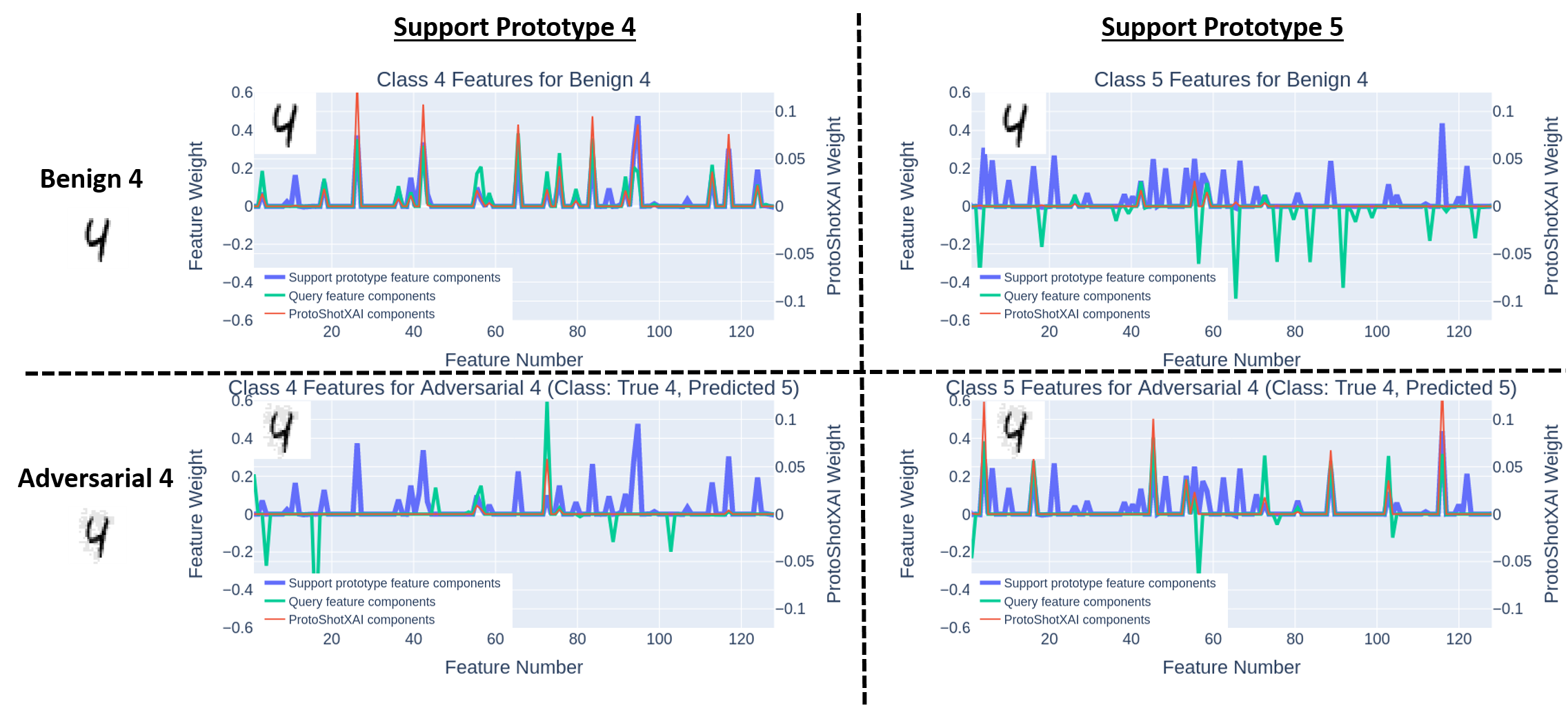}
\caption{Results on the adversarial MNIST dataset which shows the decomposition of feature components from a benign \textbf{four} (top) and adversarial \textbf{four} (bottom) w.r.t. class \textbf{four} prototype (left) and class \textbf{five} prototype (right). \rebuttaltext{The components for the support set are the values of the vector $\pbf_{S,c}$ (as defined by equation \eqref{eq:Prototype_vector}), the query set are the values of the vector $\fbf({\xbf}_{m}') \odot \wbf^{c}$, and the ProtoShotXAI components are the values of the vector $\pbf_{S,c} \odot \left[\fbf({\xbf}_{m}') \odot \wbf^{c} \right]$ (as defined in numerator of equation \ref{eq:ProtoShotXAI_vector} prior to the summation of components)}. From these plots, the benign \textbf{four} has a very positive response to the class \textbf{four} prototype and a negative response to the class \textbf{five} prototype. The adversarial samples do not show this same behavior, it is more uniformly distributed across features. ProtoShotXAI can demonstrate at this level that the features are corrupted, but what components caused feature corruption is still unanswered.}
\label{fig:adv_example}
\end{figure}

\begin{figure}
\centering 
\includegraphics[width=1.0\textwidth]{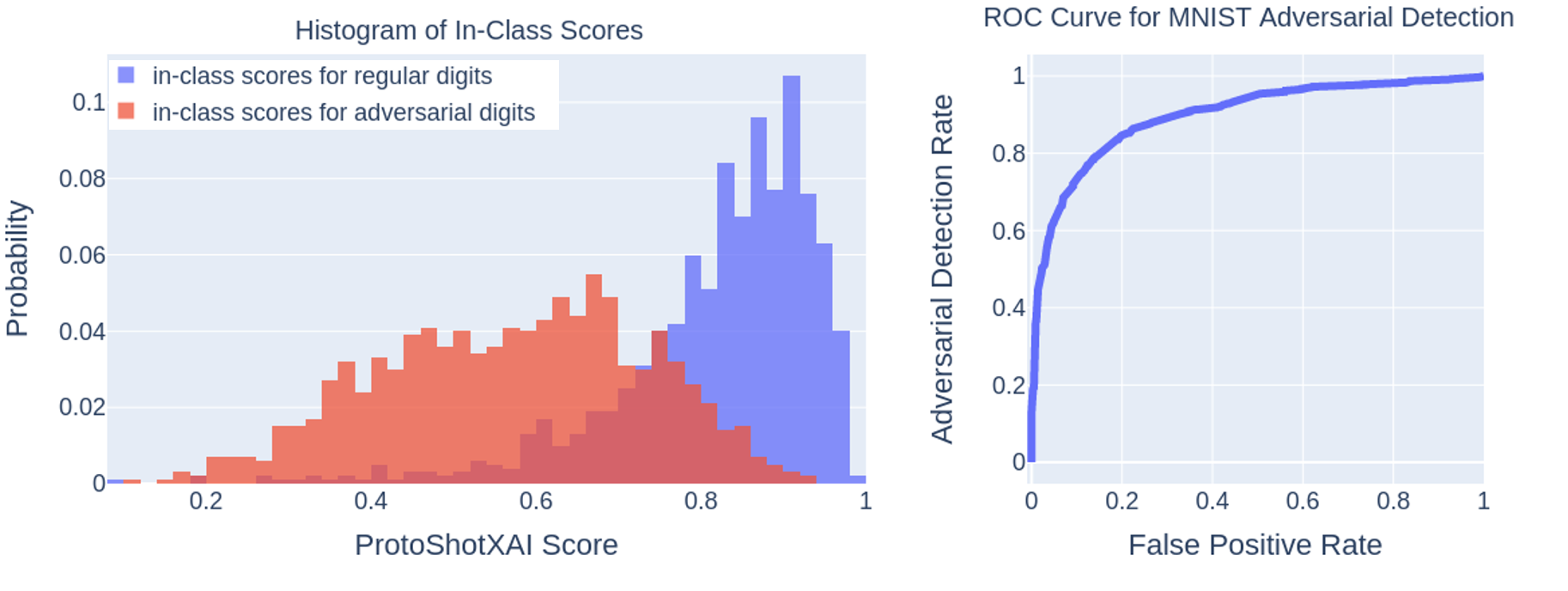}
\caption{The plot on the left shows the histogram of the in-class scores for the benign \textbf{four} (in blue), the adversarial \textbf{four} (in orange), and the overlap (in red). The plot on the right is the ROC curve for a simple histogram threshold detector applied to the ProtoShotXAI scores. The result is the ProtoShotXAI scores can be used to detect adversarial samples by knowing the in-class scores from a validation dataset.}
\label{fig:adv_hist_and_roc}
\end{figure}

The results of this experiment are shown in Figures \ref{fig:adv_example} and \ref{fig:adv_hist_and_roc}. 
There are several observations that we make by considering the results in Figure \ref{fig:adv_example}. The feature attribution maps of all methods on the adversarial data were noisy and provided little-to-no insight; however, by plotting the component weights of the support prototype, the query sample, and the ProtoShotXAI score, we noticed a clear difference between the adversarial samples and benign samples. \rebuttaltext{The components are the raw element values that make a vector. For example, in Figure \ref{fig:adv_example} the components for the support set are the values of the vector $\pbf_{S,c}$ (as defined by equation \ref{eq:Prototype_vector}), the query set are the values of the vector $\fbf({\xbf}_{m}') \odot \wbf^{c}$, and the ProtoShotXAI components are the values of the vector $\pbf_{S,c} \odot \left[\fbf({\xbf}_{m}') \odot \wbf^{c} \right]$ (as defined in numerator of equation \ref{eq:ProtoShotXAI_vector} prior to the summation of components).} 

In this experiment, a benign \textbf{four} and adversarial \textbf{four} were selected from the testing dataset to demonstrate this difference. 
As shown in the top two plots of Figure \ref{fig:adv_example}, a benign \textbf{four} correlates well to the support prototype of a \textbf{four} (top left plot), which causes prominent peaks in the ProtoShotXAI components and subsequent scores. 
The prototype of a \textbf{five} causes the query \textbf{four} to produce a lot of negative weights and is not correlated well to the support prototype of a \textbf{five} (top right plot). The result is as expected: the benign \textbf{four}'s score for the class \textbf{four} prototype is much larger than the class \textbf{five} prototype. However, in the adversarial query sample of a \textbf{four}, neither the score for the prototype of the \textbf{four} nor the prototype of the \textbf{five} are well correlated (bottom left and right plots, respectively). The overall score is poor for both classes but is marginally worse for class \textbf{four} than for class \textbf{five}. As a result, the adversarial sample is incorrectly classified as the \textbf{five} class.

The variation between the adversarial and benign scores for the in-class samples in Figure \ref{fig:adv_example} is novel to the ProtoShotXAI approach and begs the question: {\em can ProtoShotXAI be useful to identify adversarial samples?}
From the analysis of ProtoShotXAI scores, the results show that the adversarial samples corrupt the network at the feature layer by decorrelating the feature weights from the classification weights. The features in a benign sample strongly align to the weights of its own class, whereas the adversarial features do not strongly align with any class. This lack of alignment to a class causes minor correlations to decide the prediction. 
This result is made clear by running an experiment that generates the in-class ProtoShotXAI scores for 1,000 random samples of both the benign and adversarial MNIST data. The benign and adversarial score distributions are shown in the left plot of Figure \ref{fig:adv_hist_and_roc}. As expected, the adversarial digits have a significantly lower average than the average ProtoShotXAI scores on benign digits. It is also worth noting that the variance of ProtoShotXAI's score is larger than its benign counterpart. A simple detector can be created by applying a threshold to the ProtoShotXAI score. The right plot of Figure \ref{fig:adv_hist_and_roc} shows the receiver operating characteristic (ROC) curve for adversarial detection versus falsely detecting benign digits as adversaries. The results of the simple threshold detector demonstrate that ProtoShotXAI can answer when a sample is adversarial and helps to improve model trust; however, it does not directly answer why the sample is adversarial (i.e., why from an interpretability perspective). We leave the analysis of ``why'' the sample is adversarial as future work and this question of ``why'' is an ongoing area of research \citep{merrer2020nature, Tomsett2020}. 

\subsection{Omniglot}

\begin{figure}
     \centering
     \begin{subfigure}[b]{0.25\textwidth}
         \centering
         \includegraphics[width=\textwidth]{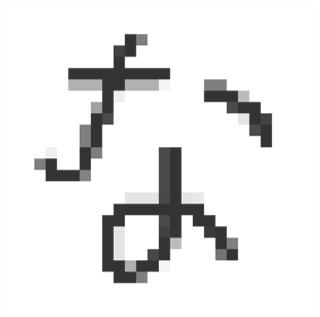}
         \caption{Omniglot character}
         \label{fig:Omniglot_Character}
     \end{subfigure}
     \hspace{1em}
     \begin{subfigure}[b]{0.25\textwidth}
         \centering
         \includegraphics[width=\textwidth]{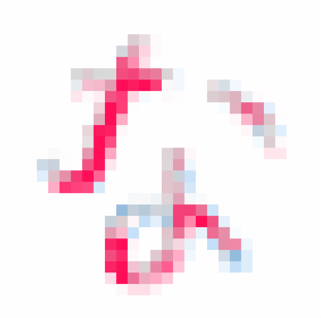}
         \caption{Feature explanations}
         \label{fig:Omniglot_Character_Features}
     \end{subfigure}
        \caption{An example of an Omniglot character (left) and ProtoShotXAI feature attributions w.r.t in-class prototype using Algorithm \ref{alg:ProtoShotXAI_Implementation} (right). The features indicate the strongest weighting to the sub-character on the top left and the weakest weighting to the sub-character on the top right. We demonstrate a similar conclusion when looking at the similarity score w.r.t removing the larger sub-characters as shown in Figure \ref{fig:omniglot_feature_removal}.}
        \label{fig:omniglot_example}
\end{figure}

ProtoShotXAI borrows from a few-shot architecture which makes it easy to apply to few-shot classification methods. To the best of our knowledge, ProtoShotXAI is the first method to explicitly apply XAI to datasets such as Omniglot. Since many few-shot architectures use the dual network structure for support and query samples, applying ProtoShotXAI to few-shot tasks is equivalent to a standard neural network with the classification weights set to a vector of ones. 
Figure \ref{fig:omniglot_example} shows a character from the Omniglot database (left) that can be classified with the trained model from Table \ref{tab:omniglot-model-summary}. Similar to MNIST, we used ProtoShotXAI with Algorithm \ref{alg:ProtoShotXAI_Implementation} and a reference pixel value of zero which indicates a lack of digit presence in a pixel. Although many pixels have a positive in-class response for the Omniglot character (right), it is observed that the order of importance appears to be the top left, the bottom, and the top right sub-character. Additionally, the specific size of the top right sub-character appears to be less important than the spatial center based on the strong positive response from the sub-character's spatial center. 

\begin{figure}
\centering 
\includegraphics[width=1.0\textwidth]{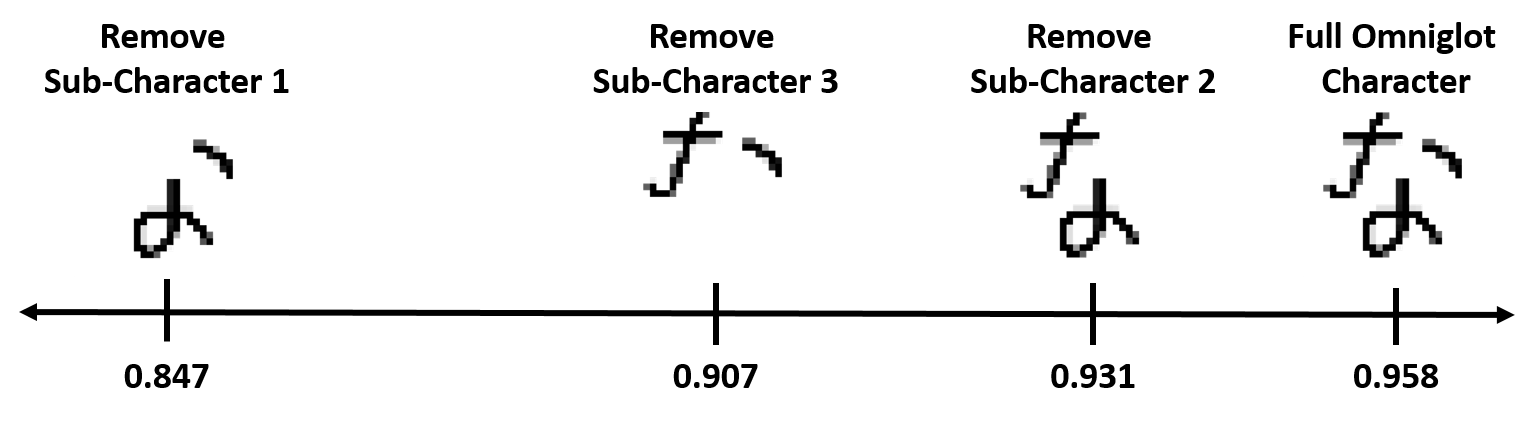}
\caption{ProtoShotXAI similarity score w.r.t in-class prototype. The image on the far right with the highest similarity score corresponds to the full Omniglot character where the three images on the left correspond to the ProtoShotXAI score with different sub-characters removed. Removing the top left sub-character causes the largest negative deviation from the similarity score, indicating it is the most important sub-character of the three. The order of sub-character importance agrees with the feature attribution maps shown in Figure \ref{fig:omniglot_example}.}
\label{fig:omniglot_feature_removal}
\end{figure}

In contrast to feature attributions, ProtoShotXAI can evaluate the response to large-scale feature changes. Using the previous character as an example, we removed each of the respective sub-characters to determine the overall network response to its class using the ProtoShotXAI score. Figure \ref{fig:omniglot_feature_removal} shows the resultant scores on a line plot. The overall score dropped the most by removing the top left sub-character, indicating that it was most important to the overall character classification. Conversely, removing the top right sub-character caused little change to the overall score, indicating that it was least important to the overall character classification. This result corresponds to the same order of sub-character importance obtained from the feature attribution maps.

\subsection{ImageNet}

\begin{figure}
\centering 
\includegraphics[width=0.7\textwidth]{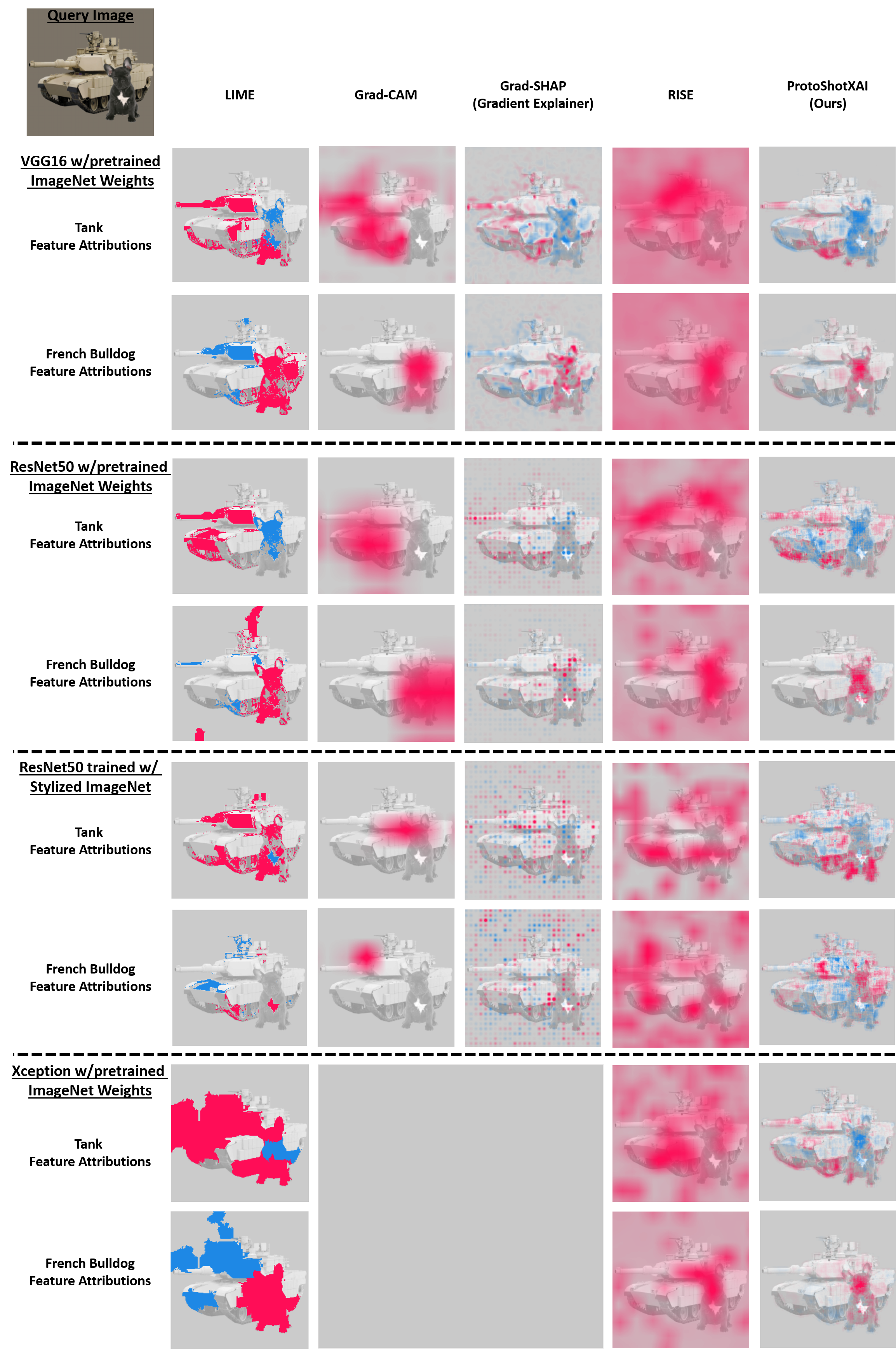}
\caption{Comparison of feature attributions produced by five different methods (LIME, Grad-CAM, Grad-SHAP, RISE, and ProtoShotXAI) and four different networks. The query image containing the tank and french bulldog is shown in the top left. Each row corresponds to a feature attribution of a given network for a specific class (i.e., tank or french bulldog class). Each column corresponds to a particular approach. Due to the complexity of the Xception Network architecture, the gradient methods, which expect a specific convolutional architecture, were not applicable. The feature attributions from ProtoShotXAI are noticeably more consistent (i.e., positive and negative feature attributions are more isolated to their respective class region).}
\label{fig:ImageNet_feature_attribution}
\end{figure}

\begin{figure}
\centering 
\includegraphics[width=0.80\textwidth]{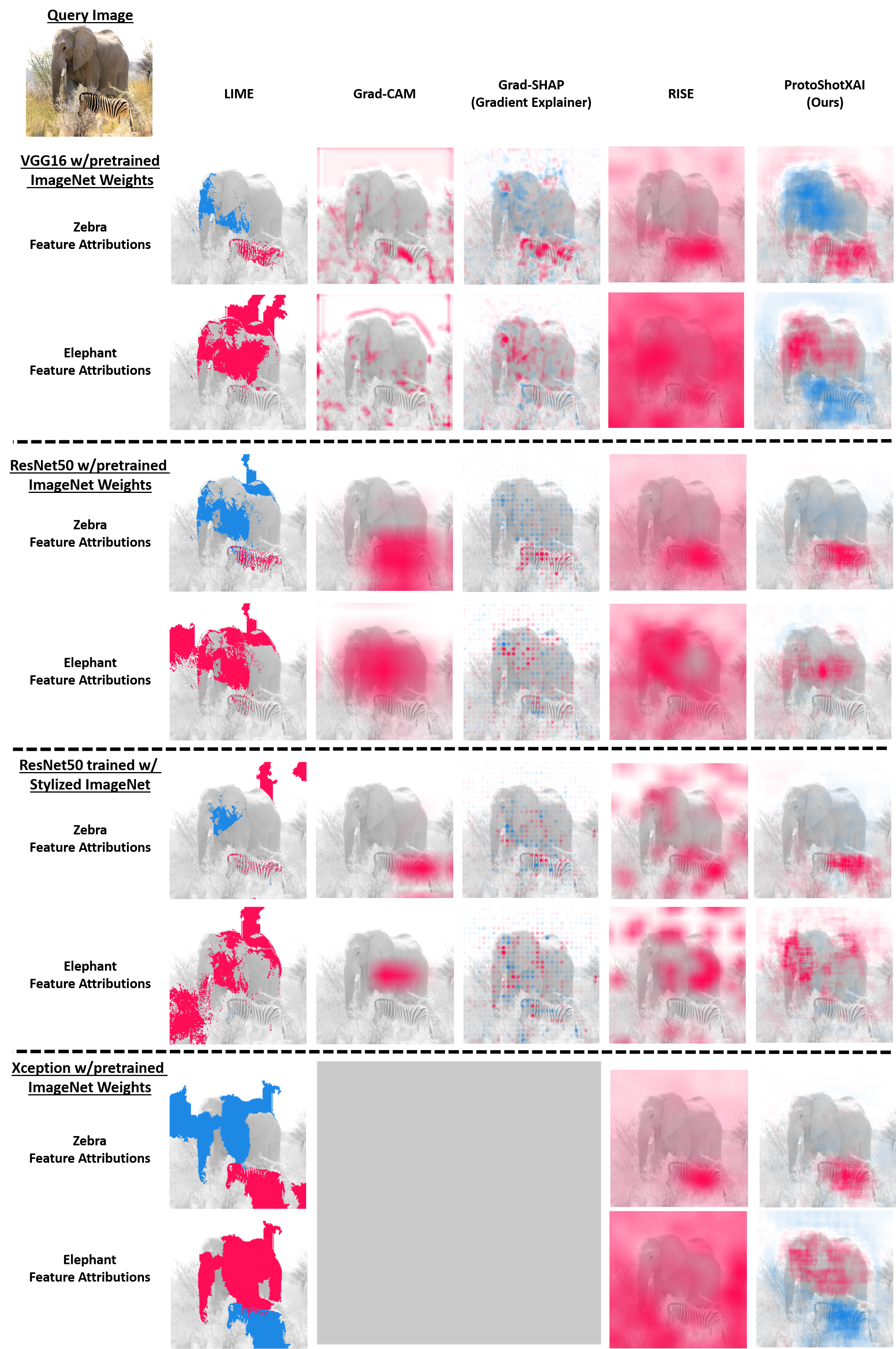}
\caption{\rebuttaltext{Comparison of feature attributions on a real image of an African elephant and a zebra. The query example is shown in the top left. Similar to the findings in Figure \ref{fig:ImageNet_feature_attribution}, ProtoShotXAI attributions are noticeably more consistent (i.e., positive and negative feature attributions are more isolated to their respective class region).}}
\label{fig:ImageNet_feature_attribution_el_zebra}
\end{figure}

\begin{figure}
\centering 
\includegraphics[width=0.80\textwidth]{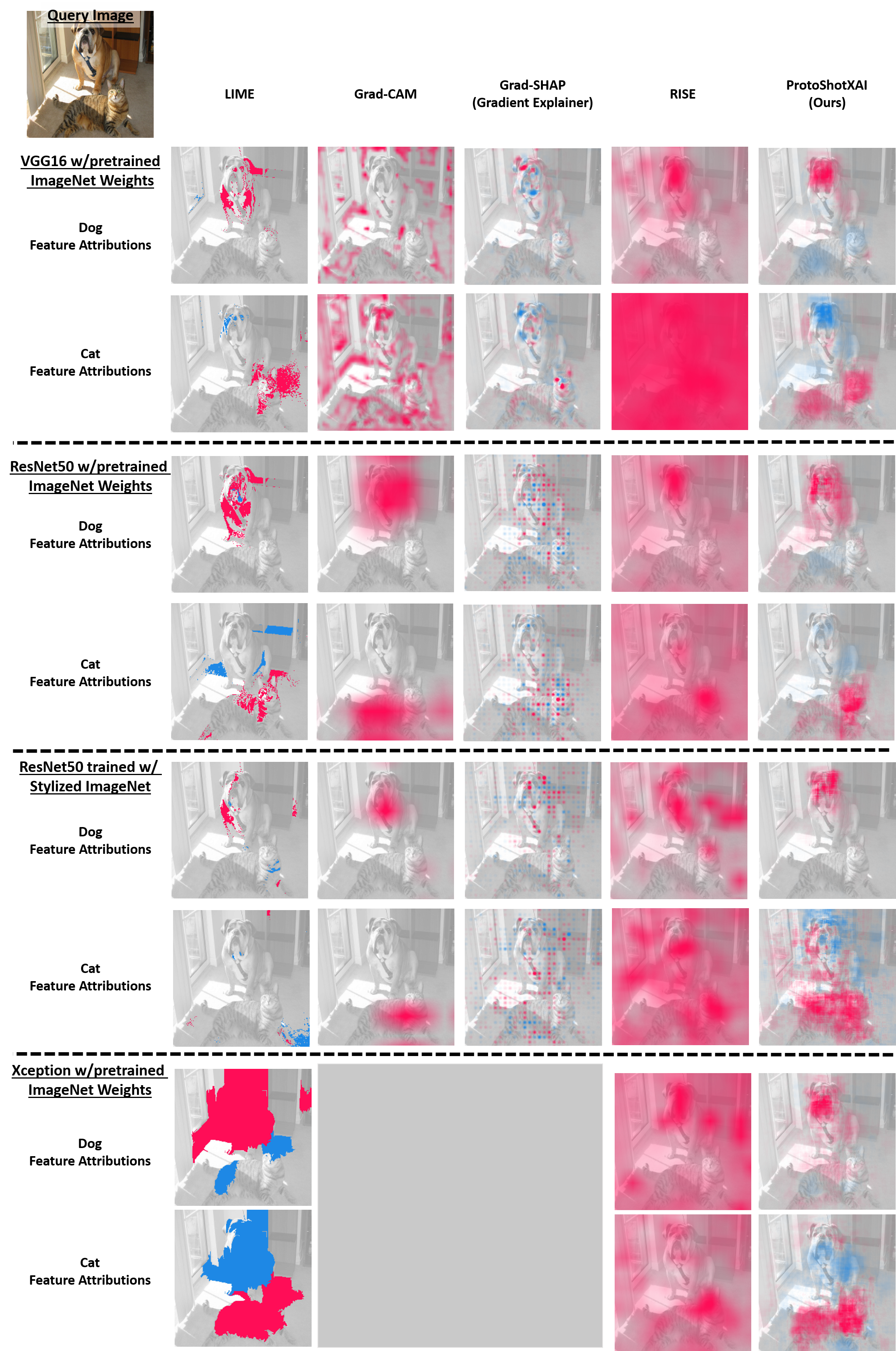}
\caption{\rebuttaltext{Another comparison of feature attributions on a real image. The query example of a cat and dog is shown in the top left. Similar to the findings in Figure \ref{fig:ImageNet_feature_attribution} and \ref{fig:ImageNet_feature_attribution_el_zebra}, ProtoShotXAI attributions are noticeably more consistent (i.e., positive and negative feature attributions are more isolated to their respective class region).}}
\label{fig:ImageNet_feature_attribution_cat_dog}
\end{figure}

The previous sections evaluated ProtoShotXAI against state-of-the-art digit classification methods (MNIST) and also demonstrated that ProtoShotXAI could be used for few-shot character classification (Omniglot). Both the MNIST and Omniglot datasets are images with approximately binary pixels; however, many real-world applications have complex features or pixels. The experiments in this section aim to qualitatively demonstrate that the ProtoShotXAI approach can be applied to datasets with complex image features, namely ImageNet. \rebuttaltext{In the first example (see Figure \ref{fig:ImageNet_feature_attribution}), we reduce contextual clues by creating an artificial input of an image containing two unrelated classes within the 1,000 ImageNet classes: a tank and a french bulldog. The background for this image is an average pixel value that appears as a brownish color and provides no additional contextual information. In Figure \ref{fig:ImageNet_feature_attribution_el_zebra} and \ref{fig:ImageNet_feature_attribution_cat_dog}, we show two more examples on real images that contain both contextual background and two class objects in a single image. Figure \ref{fig:ImageNet_feature_attribution_el_zebra} contains an elephant and a zebra, whereas Figure \ref{fig:ImageNet_feature_attribution_cat_dog} contains a cat and a dog.} For each image, we show the attributions produced from a total of four pretrained classification networks used in the assessment: VGG16, Xception Network, and ResNet50, all with ImageNets weights, and ResNet50 with Stylized-ImageNet weights. ``ImageNet weights'' refer to the default weights provided by Keras and the ``Stylized-Imagenet weights'' refer to weights from Stylized-Imagenet training as discussed in Appendix \ref{app:Implementation_Details}. Stylized-ImageNet substitutes textures in the images with the goal of training a network that is biased towards shape instead of texture.
All three figures shows the comparison of ProtoShotXAI, Grad-SHAP (SHAP Gradient Explainer), Grad-CAM, LIME, and RISE for each trained network. Only LIME, RISE, and ProtoShotXAI were applicable on Xception Network due to the network's complexity.
The gradient-based methods (i.e., Grad-SHAP and Grad-CAM) operate at a specific convolutional layer and break down with unconventional convolutional architectures (like that of Xception Network). In contrast, LIME and ProtoShotXAI use the full model and do not have the same limitation. 
In this work, Grad-CAM operates at the last convolutional layer and Grad-SHAP operates on layers 7 and 38 for the VGG16 and ResNet architectures, respectively.
The choice of the convolutional layer is a hyperparameter in Grad-SHAP where deeper layers tend to provide a better representation at the cost of lower spatial resolution. The layer was selected based on a comparable spatial resolution to the other methods, which results in a fair comparison.

For each trained network and each example image, ProtoShotXAI demonstrates consistent positive features (red) on the class of interest negative features (blue) on the counter class. For example, ProtoShotXAI demonstrates consistent positive features for the tank attributions around barrel/treads and negative features in the region of the french bulldog (see Figure \ref{fig:ImageNet_feature_attribution}). For the french bulldog attributions, ProtoShotXAI has positive responses around the face and negative responses scattered throughout the tank. All methods appear to have inconsistencies w.r.t attribution and class (e.g., some positive feature attributions on the tank for the french bulldog class); however, the response of ProtoShotXAI is more spatially consistent throughout. Grad-SHAP is the closest to ProtoShotXAI for the VGG16 class with more ``incorrect'' positive features on the tank for the french bulldog class. Additionally, Expected Gradient's dot-like appearance on ResNet50 is from the method's use of deeper convolutional layers that partition the image into low-resolution spatial features. In contrast, the other gradient method, Grad-CAM, has a more blurred appearance and only identifies positive features of interest. The blurred appearance causes background regions to be positively indicated as feature attributions as well.  
LIME does not have the same issue of positively indicated feature attributions due to a segmentation algorithm that causes very defined region edges; however, LIME also segments large regions inconsistently, such as the positive feature attributions around the french bulldog's feet w.r.t the tank class.

ResNet50 with Stylized-ImageNet is unique to this experiment because it is the only network trained with the modified ImageNet dataset meant to remove the network's texture bias while maintaining the shape bias. We expected to see different network responses that indicate isolation of feature boundaries rather than regions, and unfortunately, this behavior was not explicitly observed. The feature attribution maps for ResNet50 trained with Stylized-ImageNet are drastically different from ResNet50 trained with the original ImageNet but it is inconclusive to argue that any XAI feature attribution methods show evidence of a shape bias. Although considered out of scope for this work, our XAI results warrant further investigation to determine if ResNet50 (trained with Stylized-ImageNet) has learned some complex features other than explicit shapes, such as the underlying textures from the stylized transformation. 

\section{Discussion}
\label{Discussion}

Throughout the experiments section, ProtoShotXAI was demonstrated to be a flexible model exploration tool that produces comparable (and often superior) qualitative and quantitative results.
ProtoShotXAI is uniquely straightforward to implement since it does not require training another model (e.g., LIME), or a CNN architecture (e.g., Grad-SHAP and Grad-CAM). ProtoShotXAI only requires a network's feature layer and data from the classes to construct prototypes. 
Further, comparable feature attribution maps are created using a simple pixel perturbation which could be enhanced with more intelligent perturbation methods. 

Although not explicitly presented in this work, we also attempted different variations of ProtoShotXAI. First, we found that the classification weights are needed to produce negative feature attributions because the output of the feature layer often consists of ReLU activations in state-of-the-art neural networks. ReLU activations are either positive or zero, meaning the cosine distance at the feature layer without the classification weights will not produce strong negative feature attributions. In contrast, the Euclidean distance can be used as the similarity score; however, the normalization of cosine distance provided better interpretability across features and was more consistent with the original model response in experiments such as the revolving {six}. Additionally, the distance between the prototype and query sample can be computed at the output of the final classification layer rather than the weighted feature layer. We found this also produced less favorable results. More specifically, it is common for the final classification layer to have lower dimensionality than the feature layer and a softmax activation function, rather than an unbounded ReLU activation at the feature layer. Both the lower dimensionality and bounded activation functions limited the feature attribution interpretability at the classification layer. The class-weighted feature layer produced the best results in our experiments.
We also investigated training models to enhance our approach. For example, we implemented an augmentation to the baseline classifier with a few-shot training paradigm. The training objective was to learn adaptive weights between the prototype features and query features such that it would minimize the distance between like classes and maximize the distance between unlike classes. Although the model successfully trained the dynamic weights, it only produced marginally better results. We did not include it here because the benefit of the augmented model was outweighed by the complexity and time cost of having an extra training process. Conveniently, ProtoShotXAI, as presented in this work, does not require any additional training.

\section{Conclusions}
\label{Conclusions}

Machine learning, particularly deep learning, has been shown to provide state-of-the-art performances for numerous benchmarks; however, the explainability of these models and how they reach a decision is a challenging task that limits their utility in some applications. Consequently, it is nearly impossible to quantitatively assess ML approaches' performance to explain black-box models directly. Instead, indirect qualitative and quantitative measures are used for XAI algorithm performance assessment. Qualitative measures can produce misleading results by exploiting an observer's confirmation bias and indirect quantitative measures are imperfect assessments. As a result, there is no one-size-fits-all XAI approach and the best approach depends on the data and application. It was against this background that we proposed the ProtoShotXAI architecture.  ProtoShotXAI is an approach to aid ML practitioners in exploring their black-box model. We demonstrated on three datasets that ProtShotXAI can create feature attribution maps that are qualitatively comparable, and arguably superior, to state-of-the-art methods. This work also showed how ProtoShotXAI can be extended to higher-level feature analysis by examining the similarity scores as a function of complex changes to the input as demonstrated in the MNIST revolving \textbf{six} and Omniglot sub-character removal experiments. We demonstrate both quantitatively and qualitatively that ProtoShotXAI provides more flexibility for feature analysis and model exploration. Further, the components of the similarity score provide additional insight into model behavior, as demonstrated in the adversarial MNIST experiment. ProtoShotXAI in these specific experiments demonstrates that a user can apply ProtoShotXAI as an exploration tool that will aid in improving the trust in black-box models.

The data used in this work was instrumental in understanding how XAI can be used and evaluated in practice. The digit \textbf{eight} was selected due to the convenient presence of other digits within the trace of an \textbf{eight}. The image containing the tank and french bulldog was created as a test image without contextual clues and two, and only two, classes. All of the other data manipulations (i.e., revolving \textbf{six}, sub-character analysis in Omniglot, adversarial analysis) were done intelligently. The intelligent manipulations allowed us to make apriori assumptions about a model's outcome that could be validated against the information provided by ProtoShotXAI. Our future work aims to expand the intelligent data manipulations done in this work into more comprehensive datasets to evaluate XAI approaches further. \rebuttaltext{Specifically, quantitative metrics may be obtained by carefully occluding or removing objects from images in ways that appear visually natural, especially in images that contain multiple objects (classes) of interest. Carefully designed surveys could also help, especially if the surveys account for human-in-the-loop interaction and hypothesis testing.} Additionally, ProtoShotXAI is applicable to classification networks for many different data types. Although the scope of this work was limited to images, future work will extend ProtoShotXAI to classification networks with sound, text, and data signals.

\section*{Acknowledgements}
This work was supported by grants from the Department of Energy \#DE-NA0003946, Army Research Lab W56KGU-20-C-0002, the National Science Foundation (NSF) CAREER \#1943552, and NSF \#1950359. Any opinions, findings, and conclusions or recommendations expressed in this material are those of the authors and do not necessarily reflect the views of the sponsors.

\bibliography{ProtoShotXAI_refs,greg}

\appendix
\section{\rebuttaltext{Implementation Details}}
\label{app:Implementation_Details}

\paragraph{Datasets} 
In this work, we use three commonly used datasets to evaluate the interpretability of our approach: MNIST, Omniglot, and ImageNet. The MNIST dataset is a well-established benchmark for image classification. Although high classification performance can be achieved with very simple models, it is convenient to revisit this dataset in the context of XAI due to the overlap between features of different classes. For example, the number \textbf{eight} has a unique cross-over ``x''-like feature in the center; however, if the left half of the \textbf{eight} were removed then the remaining features appear like a \textbf{three}. Similarly, if the diagonal line from the bottom left to the top right of the \textbf{eight} were removed then the remaining features appear more like a \textbf{five}. We would expect similar conclusions when comparing XAI approaches on sufficiently trained classification networks. 

The Omniglot dataset is a character database that is similar to MNIST but is used to benchmark few-shot algorithms. The dataset is composed of 50 different alphabets, and a total of 1,623 character sets \citep{lake2011one-sam}. For each of the 1,623 characters, there are only 20 human-drawn samples which result in a database size of 32,460 samples. The data are originally 105x105 binary images and are resized to 28x28 to reduce input dimensionality. \rebuttaltext{1,200 characters are used for training/validation and the remaining set of 423 characters are used for a disjoint testing set. Within their respective disjoint sets, images are rotated in multiples of 90 degrees to produce a set of 4,800 classes (96,000 samples) for training/validation and 1,692 classes (423 with \ang{90} rotations) for testing (33,840 samples)}. To our knowledge, the Omniglot dataset has not been used in any other XAI approach and its application is unique to our work.

ImageNet is a low-resolution image database that is also known as the ILSVRC-12 database \citep{russakovsky2015imagenet}. Samples vary in size but are often reduced to an image of size 3x224x224 RGB (as in this work) or less. The entire dataset is made up of nearly 14 million samples, but the scale-invariant feature set used in computer vision problems (and in this work) contains 1,000 classes with 1,200 samples per class (i.e., 1.2 million samples total). There is also a subset of the scale-invariant ImageNet, known as Stylized-ImageNet, that we use in this work. Stylized-ImageNet substitutes textures in the images with the goal of creating a training dataset that is biased towards shape instead of texture \citep{geirhos2018imagenettrained}. For example, the fur from a dog is replaced with textures of skin from an elephant or bricks from a house while maintaining the shape and resemblance of the dog.

\paragraph{Classification Architectures} 
All the XAI approaches evaluated in this work are applied to trained classification architectures, so each dataset was paired with a suitable neural network. For the MNIST dataset, we used the convolutional network shown in Table \ref{tab:mnist-model-summary}  which has a total of 1.2 million trainable parameters. For the Omniglot dataset, we followed the configuration used in prototypical networks \citep{snell2017prototypical} and shown in Table \ref{tab:omniglot-model-summary} with 112K trainable parameters. Finally, for ImageNet, we used the VGG16, ResNet50, and Xception networks available in the Keras applications library. VGG16 was chosen as a common benchmark between methods because most comparable XAI approaches can be easily applied. VGG16 has 71\% classification accuracy and 138 million parameters. Conversely, Xception network has a relatively small number of parameters (22 million) and is one of the best performing networks available in Keras (79\% classification accuracy, third only to NASNetLarge with 89M parameters and 83\% performance and InceptionResNetV2 with 56M parameters and 80\% performance). 
Due to the complexity and depth of the Xception network (i.e., 126 layers with multiple convolutional layers, separable convolutional layers, and skip layers), only ProtoShotXAI and LIME can be easily applied to this network, and the other methods fail due to limited computational resources. Additionally, ResNet50 was used to compare conventional ImageNet training and a shape-based variant of ImageNet, Stylized ImageNet. The network architectures were equivalent for both ImageNet and Stylized ImageNet, only the weights are different. VGG16 is shown in Table \ref{tab:imagenet-model-summary}. For the specific ResNet50 and Xception network details, we refer the reader to the figures in the original papers \citep{he2016deep, chollet2017xception} or the supplemental material section on Github\footnote{\url{https://github.com/samuelhess/ProtoShotXAI/}}.

\begin{table} 
\begin{tabular}{lll} 
Model: Convolutional MNIST Model  \\ \hline 
Layer                           & Output Shape                & \# of Parameters    \\ \hline \hline
Input                                               & (28, 28, 1)           & 0       \\ \hline 
Conv2D (ReLU Activation)                            & (26, 26, 32)          & 320        \\ \hline 
Conv2D (ReLU Activation)                            & (24, 24, 64)          & 18496      \\ \hline 
MaxPooling2D                                        & (12, 12, 64)          & 0          \\ \hline 
Dropout 25\%                                        & (12, 12, 64)          & 0          \\ \hline 
Flatten                                             & (9216)                & 0          \\ \hline 
\textbf{Dense Feature Layer (ReLU Activation)}$^*$               & (128)                 & 1179776    \\ \hline 
Dropout 50\%                                        & (128)                 & 0          \\ \hline 
Dense Classification Layer (Softmax Activation)     & (10)                  & 1290       \\ \hline \hline 
Total parameters: 1,199,882 \\ 
Trainable parameters: 1,199,882 \\ 
Non-trainable parameters: 0 \\ \hline
\multicolumn{3}{l}{$^*$ The output of this layer corresponds to $\fbf(\xbf)$ in Section \ref{approach}.}
\end{tabular} 
\caption{Convolutional MNIST Model} 
\label{tab:mnist-model-summary} 
\end{table}

\begin{table} 
\begin{tabular}{lll} 
Model: Prototypical Omniglot Model \\ \hline 
Layer                                   & Output Shape                & \# of Parameters    \\ \hline \hline
Input                                   & (28, 28, 1)           & 0       \\ \hline 
Conv2D (BatchNorm + ReLU Activation)    & (28, 28, 64)          & 896        \\ \hline 
MaxPooling2D                            & (14, 14, 64)          & 0          \\ \hline 
Conv2D (BatchNorm + ReLU Activation)    & (14, 14, 64)          & 37184      \\ \hline 
MaxPooling2D                            & (7, 7, 64)            & 0          \\ \hline 
Conv2D (BatchNorm + ReLU Activation)    & (7, 7, 64)            & 37184      \\ \hline 
MaxPooling2D                            & (3, 3, 64)            & 0          \\ \hline 
Conv2D (BatchNorm + ReLU Activation)  & (3, 3, 64)              & 37184      \\ \hline 
MaxPooling2D                            & (1, 1, 64)            & 0          \\ \hline 
\textbf{Flatten (Feature Layer)}$^*$                 & (64)                  & 0          \\ \hline \hline 
Total params: 112,448 \\ 
Trainable params: 111,936 \\ 
Non-trainable params: 512 \\ \hline
\multicolumn{3}{l}{$^*$ The output of this layer corresponds to $\fbf(\xbf)$ in Section \ref{approach}.}
\end{tabular} 
\caption{Prototypical Omniglot Model} 
\label{tab:omniglot-model-summary} 
\end{table}

\begin{table} 
\begin{tabular}{lll} 
Model: VGG16 ImageNet Model \\ \hline 
Layer                                   & Output Shape                & \# of Parameters    \\ \hline \hline
Input                                   & (244, 244, 3)         & 0          \\ \hline 
Conv2D (BatchNorm + ReLU Activation)    & (224, 224, 64)        & 1792        \\ \hline
Conv2D (BatchNorm + ReLU Activation)    & (224, 224, 64)        & 36928        \\ \hline 
MaxPooling2D                            & (112, 112, 64)        & 0          \\ \hline 
Conv2D (BatchNorm + ReLU Activation)    & (112, 112, 128)       & 73856      \\ \hline
Conv2D (BatchNorm + ReLU Activation)    & (112, 112, 128)       & 147584      \\ \hline 
MaxPooling2D                            & (56, 56, 128)         & 0          \\ \hline 
Conv2D (BatchNorm + ReLU Activation)    & (56, 56, 256)         & 295168      \\ \hline
Conv2D (BatchNorm + ReLU Activation)    & (56, 56, 256)         & 590080      \\ \hline 
Conv2D (BatchNorm + ReLU Activation)    & (56, 56, 256)         & 590080      \\ \hline 
MaxPooling2D                            & (28, 28, 256)         & 0          \\ \hline
Conv2D (BatchNorm + ReLU Activation)    & (28, 28, 512)         & 1180160      \\ \hline
Conv2D (BatchNorm + ReLU Activation)    & (28, 28, 512)         & 2359808      \\ \hline 
Conv2D (BatchNorm + ReLU Activation)    & (28, 28, 512)         & 2359808      \\ \hline 
MaxPooling2D                            & (14, 14, 512)         & 0          \\ \hline
Conv2D (BatchNorm + ReLU Activation)    & (14, 14, 512)         & 2359808      \\ \hline
Conv2D (BatchNorm + ReLU Activation)    & (14, 14, 512)         & 2359808      \\ \hline 
Conv2D (BatchNorm + ReLU Activation)    & (14, 14, 512)         & 2359808      \\ \hline 
MaxPooling2D                            & (7, 7, 512)           & 0          \\ \hline
Flatten                                 & (25088)               & 0          \\ \hline
Fully Connected Layer                   & (4096)                & 102764544   \\ \hline
\textbf{Fully Connected Feature Layer}$^*$           & (4096)                & 16781312   \\ \hline
Dense Classification Layer (Softmax Activation)             & (1000)                & 4097000     \\ \hline \hline 
Total params: 138,357,544 \\ 
Trainable params: 138,357,544 \\ 
Non-trainable params: 0 \\ \hline
\multicolumn{3}{l}{$^*$ The output of this layer corresponds to $\fbf(\xbf)$ in Section \ref{approach}.}
\end{tabular} 
\caption{VGG16 ImagNet Model} 
\label{tab:imagenet-model-summary} 
\end{table}

\paragraph{Classification Training, Optimization, and Performance}
To train the MNIST and Omniglot networks, we used the ADAM optimizer \citep{kingma2014adam-sam}. The learning rate for the MNIST and Omniglot network was initially set to $6\times10^{-5}$ and $1\times10^{-3}$, respectively. A scheduler was used for the Omniglot experiments to decrease the learning rate by half every 100 epochs for a total of 1,000 epochs. We did not use a scheduler in the MNIST network because a sufficient performance was achieved with a constant learning rate after 100 epochs. For the VGG16, ResNet50, and Xception ImageNet, we used the available pretrained weights with the exception of one experiment where we trained the ResNet50 model from scratch using Stylized-ImageNet data intended to remove texture bias. For the ResNet50 Stylized-ImageNet training, the SGD optimizer was used with momentum. The initial learning rate was set to 0.1 and decreases by 90\% every 20 epochs for a total of 40 epochs. \rebuttaltext{For reference, the accuracy of each trained model on their respective dataset are shown in Table \ref{tab:Base_model_performance}. Accuracy is reported as the top-1 classification accuracy with the exception of the few-shot network which is reported as the 1-shot, 20-way, classification. As expected, the accuracy on the Stylized-ImageNet dataset is notably lower due to the applied texture randomization.}

\begin{table}
  \centering
  \begin{tabular}{l c c }
    \toprule
    \textbf{Model} &  \textbf{Dataset} & \textbf{Top-1 Classification Accuracy}\\
    \toprule
Prototypical Convolutional Omniglot$^*$ & Omniglot	& 98.1\% \\    
Convolutional MNIST 	 & MNIST   & 99.2\%\\ 
VGG16	& ImageNet		& 71.3\%  \\
ResNet50		& ImageNet		& 74.9\% \\
Xception	& ImageNet	& 79.0\% \\
ResNet50		& Stylized-ImageNet		& 43.4\%  \\ \toprule
$^*$1-shot, 20-way classification.
  \end{tabular}
   \caption{\rebuttaltext{Test classification accuracy for each model used in the evaluation of ProtoShotXAI}}\label{tab:Base_model_performance}
\end{table}

\paragraph{XAI Experimental Protocol}
The experimental protocol was to train each network with their respective data to achieve sufficient performance. After the network was trained, we constructed the ProtoShotXAI architecture presented in Section \ref{Architecture} and demonstrated in Figure \ref{fig:NetworkLayout}. We used the feature layer presented in bold for each network in each of the respective tables. When available, we applied the weights of the classification nodes to the features for the respective class. For comparison, each approach is used to explain the same image with the same model. Many approaches display feature attribution weights differently, but we used the approach developed by SHAP for all methods to maintain consistency for comparison. That is, once the feature attribution weights $\mathbf{z}$ are computed, we compute the 99.9 percentile of the absolute values of $\mathbf{z}$. The negative and positive of this value represents the dynamic range of the color axis where red and blue colors indicate positive and negative feature attributions, respectively.

\section{\rebuttaltext{ProtoShotXAI Hyperparameter Study}}
\label{app:Parameters}
\rebuttaltext{ProtoshotXAI has three hyperparameter settings for attribution map generation: (1) the number of support samples, (2) the reference value used for pixel perturbation, and (3) the size of the perturbation area. In this section, we independently vary each parameter to demonstrate the effect the parameter has on various images. We have chosen a general set of hyperparameters that produce reasonable results w.r.t. qualitative analysis, which is consistent with several other XAI approaches.}

\begin{figure}
\centering 
\includegraphics[width=1.0\textwidth]{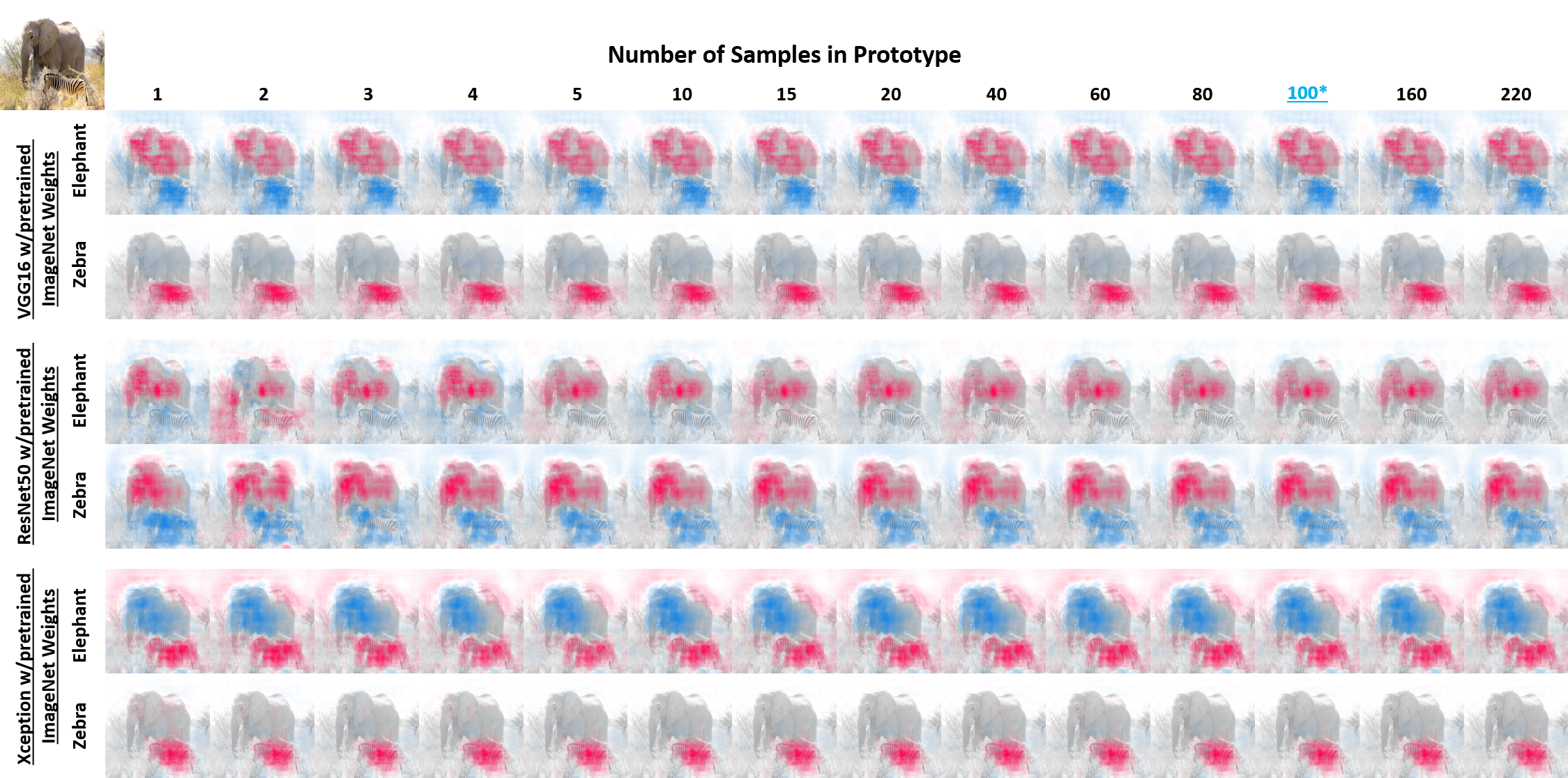}
\caption{\rebuttaltext{ProtoShotXAI feature attributions for an African elephant and zebra using VGG16, ResNet50, and Xception network classifiers. From left to right, the number of samples included in the reference support set is varied from 1 to 220. More variance is seen in the lower number of samples and all the attributions appear to converge by 60 sample support set. 100 sample support set was used throughout our work.}}
\label{fig:prototype_samples}
\end{figure}

\paragraph{\rebuttaltext{Number of support samples}}
\rebuttaltext{ProtoShotXAI is designed for sample-to-sample comparisons, or comparisons between a query image against an entire class. In the latter case, an entire class is approximated via a prototypical representation of the set of support samples. 
As observed in the few-shot literature, convergence to the true class prototype is generally achieved by averaging more samples when creating the prototype. 
Our first hyperparameter experiment demonstrates the ProtoShotXAI attributions w.r.t the number of samples in the support set. Improved stability can be visually observed as the change in attributions from subsequent images decreases.
The results are shown in Figure \ref{fig:prototype_samples}. The number of samples varies from 1 to 220 in this experiment and are selected uniformly at random without replacement from the support set. As predicted, the variation in the attribution result is larger when the number of samples is lower. It is observed that the difference between the attribution images becomes negligible as the number of samples approaches 60. This observation is consistent across several different images we investigated (see Figure \ref{fig:prototype_samples}). 
We used 100 samples in Section \ref{Experiments}'s experiments a hyperparameter in our work, ensuring consistent results w.r.t a class support set.} 

\begin{figure}
\centering 
\includegraphics[width=0.8\textwidth]{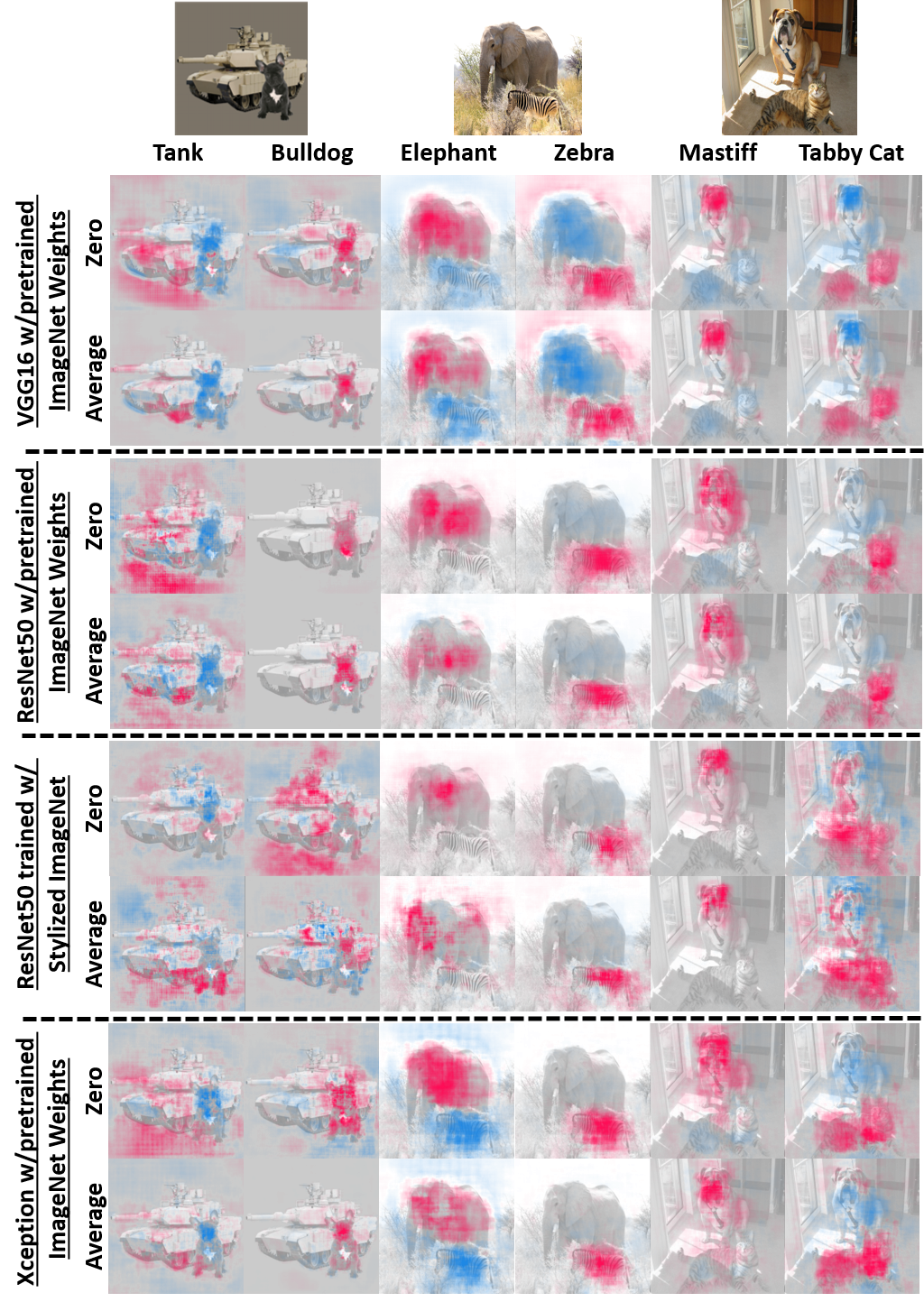}
\caption{\rebuttaltext{ProtoShotXAI feature attributions for three images using VGG16, ResNet50, and Xception network classifiers. For each image and each network, both the zero reference and average value reference for pixel perturbation are shown. Results are visually similar between the two approaches but the zero reference method has more attribution weight magnitude to the background in the tank/bulldog image.}}
\label{fig:ref_pixel}
\end{figure}

\paragraph{\rebuttaltext{Reference Perturbation Value}}
\rebuttaltext{
Our second hyperparameter experiment examines the value set for the perturbation pixel in Algorithm \ref{alg:ProtoShotXAI_Implementation}. There are several approaches that could be used to set the perturbed pixel value. Such approaches could include setting the perturbation to a zero reference (i.e., black color), maximum reference (i.e., white color), average value (i.e., brownish color), any value on $[0,1]$, or even a complex perturbation that is a function of the pixel value. Figure \ref{fig:ref_pixel} shows the ProtoShotXAI feature attributions using zero and an average value references. The attributions are very similar between the two methods; however, the zero reference method tends to have larger attribution weights across the image. The larger weights are most visible in the background attributions on the tank/bulldog image. We suspect this behavior of the zero reference is due to the fact that the perturbation value is at the extreme of pixel values. Thus, the result will likely cause more variation in the feature manifold of the network.
} 

\begin{figure}
\centering 
\includegraphics[width=1.0\textwidth]{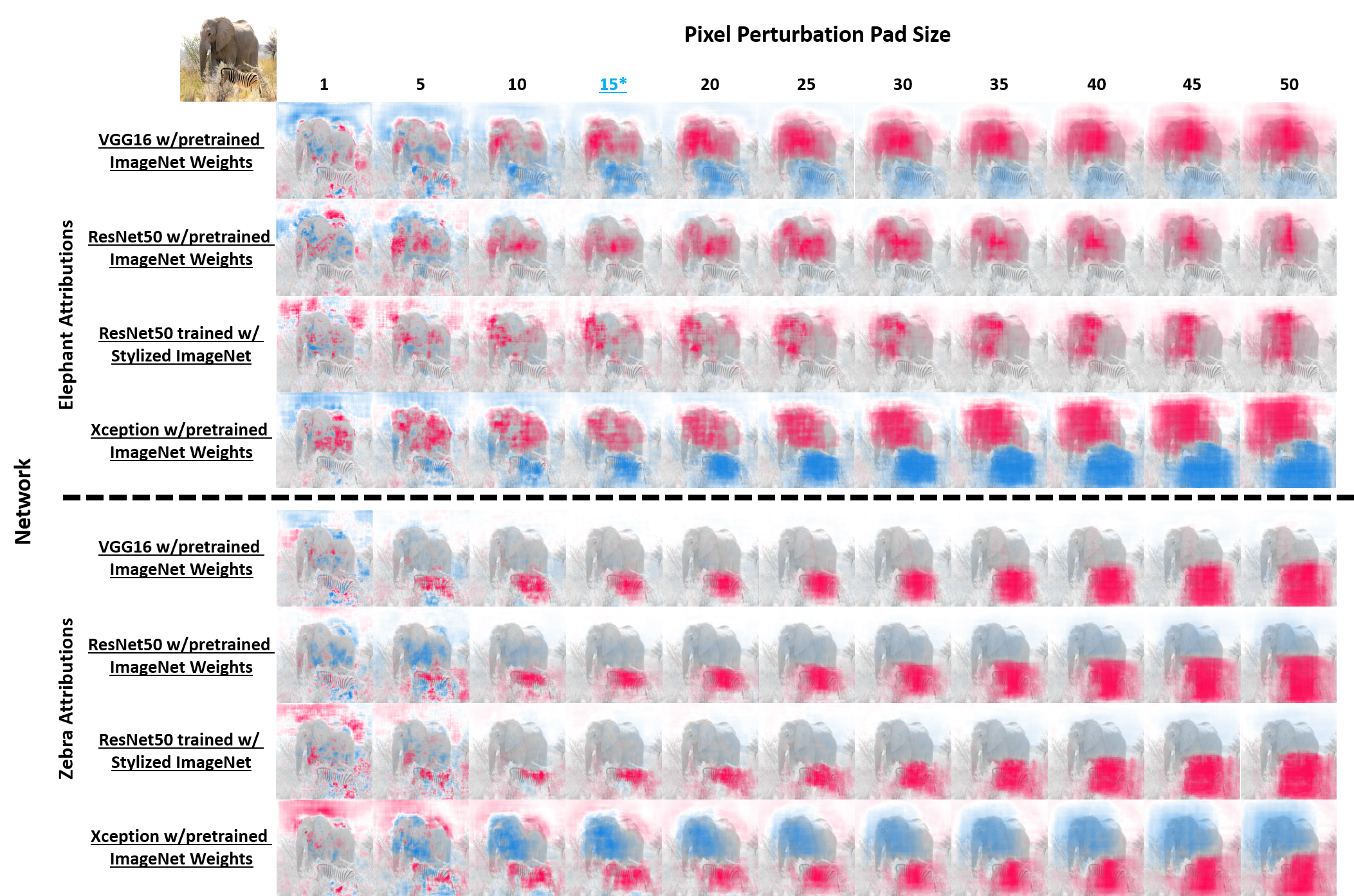}
\caption{\rebuttaltext{Comparison of ProtoShotXAI feature attributions w.r.t different perturbation areas for various networks trained for ImageNet classification. The top half of the figure demonstrates attributions w.r.t. an African elephant classification and the bottom half is w.r.t. zebra classification. There is a trade off between the pad size. As the pad size increases, the attributions cover more of the object under evaluation but the detail of the attribution is reduced. Unless specified otherwise, the pixel pad of 15 was used throughout our work.}}
\label{fig:pixel_perturbation}
\end{figure}

\paragraph{\rebuttaltext{Size of Perturbation Area}}
\rebuttaltext{Our third hyperparameter experiment varies the size of the perturbation area. The perturbation area is defined by a square pixel region centered around a pixel of interest. The entire area is replaced by the reference pixel to influence larger deviations in the feature manifold w.r.t. a spatial area of the image. 
Each pixel in the image is evaluated as a pixel of interest so the perturbation is equivalent to a moving filter and has a spatial averaging-like effect. 
The pad represents the number of pixels that surrounds the center pixel of interest in both vertical and horizontal spatial dimensions. Figure \ref{fig:pixel_perturbation} shows the results of increasing the pad size between 1 and 50 for the four networks trained on ImageNet. The support set was set to 100 samples and the average value was used for perturbation. As the pad size increases, the ProtoShotXAI attributions cover more of the object under evaluation (i.e., the elephant or zerbra); however, the detail of the attribution is reduced. For networks based around the ImageNet classification of 224$\times$224 RGB images we found that a pixel pad of 15 works qualitatively well.}

\end{document}